\documentclass[lettersize,journal]{IEEEtran}
\usepackage{amsmath,amsfonts}
\usepackage{algorithm}
\usepackage{algorithmicx}
\usepackage{algpseudocode}
\floatname{algorithm}{Algorithm} %算法
\usepackage{array}
\usepackage[caption=false,font=normalsize,labelfont=sf,textfont=sf]{subfig}
\usepackage{textcomp}
\usepackage{stfloats}
\usepackage{url}
\usepackage{hyperref}
\usepackage{verbatim}
\usepackage{graphicx}
\usepackage{cite}

\begin{document}

\title{Discrete Messages Improve Communication Efficiency among Isolated Intelligent Agents}
\author{Hang Chen, Yuchuan Jiang, Weijie Zhou, Cristian Meo, Ziwei Chen, Dianbo Liu}
% \author{Yuchuan Jiang}
% \author{Weijie Zhou}
% \author{Cristian Meo}
% % \author*[1,2]{\fnm{First} \sur{Author}}\email{iauthor@gmail.com}
% % \author[1]{\fnm{Ziwei} \sur{Chen}\textsuperscript{\Letter}}
% \author{Ziwei Chen}
% \author{Dianbo Liu}
% \email{zwchen@bjtu.edu.cn }
% \email{dianbo@nus.edu.sg}
        % <-this % stops a space



\maketitle

\begin{abstract}
Individuals, despite having varied life experiences and learning processes, can communicate effectively through languages. This study aims to explore the efficiency of language as a communication medium. We put forth two specific hypotheses: First, discrete messages are more effective than continuous ones when agents have diverse personal experiences. Second, communications using multiple discrete tokens are more advantageous than those using a single token. To validate these hypotheses, we designed multi-agent machine learning experiments to assess communication efficiency using various information transmission methods between speakers and listeners. Our empirical findings indicate that, in scenarios where agents are exposed to different data, communicating through sentences composed of discrete tokens offers the best inter-agent communication efficiency. The limitations of our finding include lack of systematic advantages over other more sophisticated encoder-decoder model such as variational autoencoder and lack of evluation on non-image dataset, which we will leave for future studies.
\end{abstract}


\section{Introduction}\label{sec1}
\begin{figure}[htbp]
	% \centering
        % \vspace{-0.35cm}  %调整图片与上文的垂直距离
	\begin{minipage}{0.45\linewidth}
		\centering
		\includegraphics[width=0.9\linewidth]{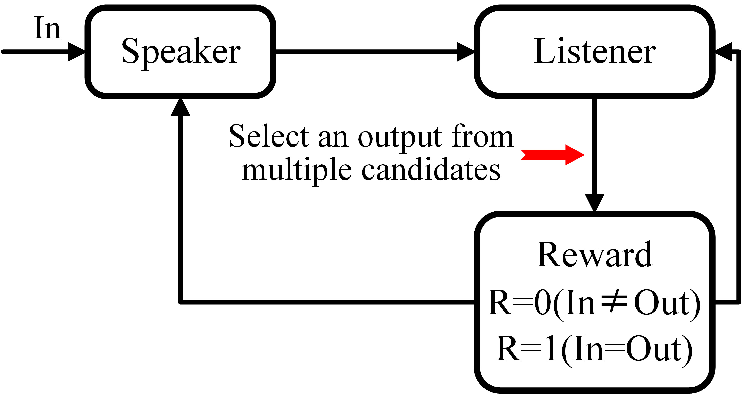}
		% \caption{图1}
		% \label{图1}%文中引用该图片代号
	\end{minipage}
        \begin{minipage}{0.48\linewidth}
		\centering
		\includegraphics[width=0.9\linewidth]{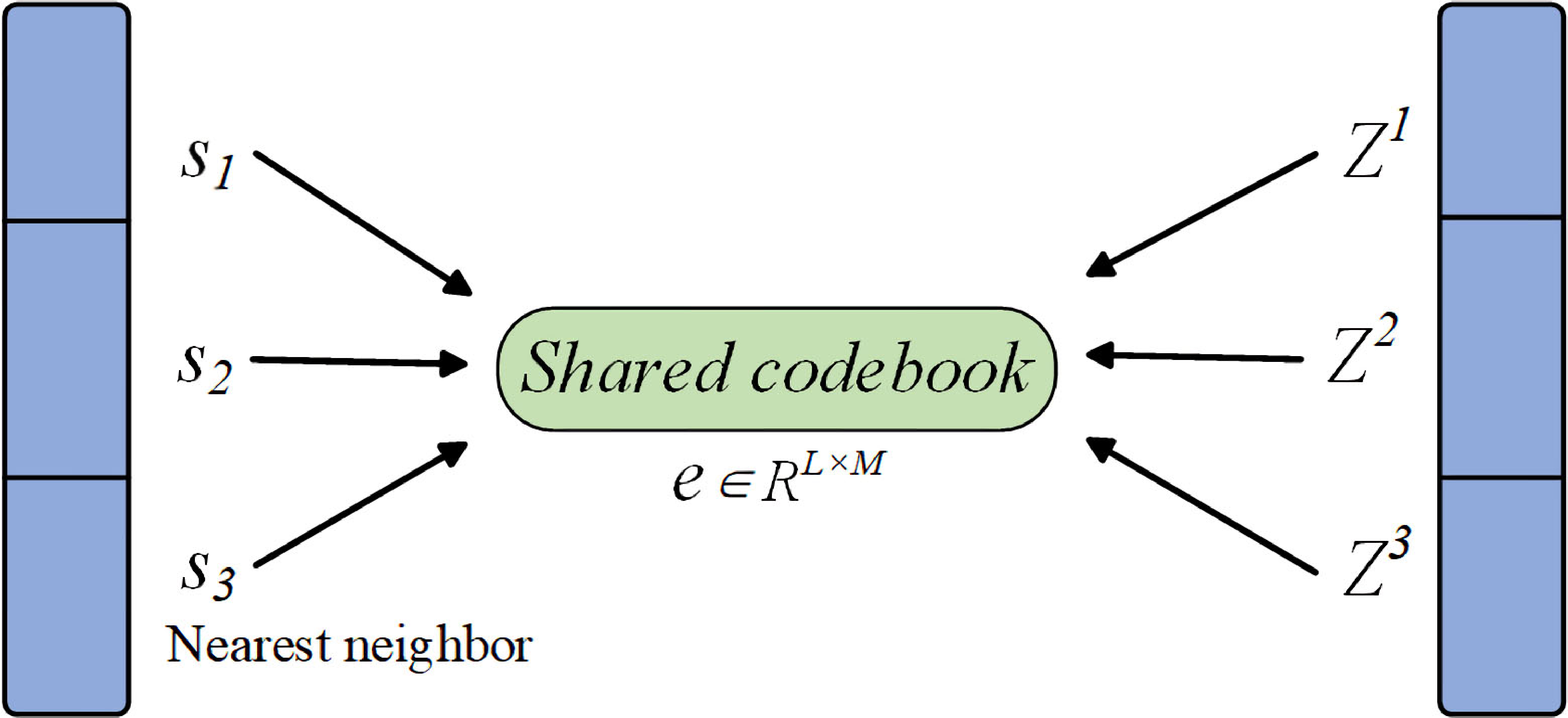}
		% \caption{图2}
		% \label{图2}%文中引用该图片代号
	\end{minipage}
        % \centering
	% \setlength{\abovecaptionskip}{0.cm} %调整标题上方的距离
        \setlength{\abovecaptionskip}{0.1cm} %调整标题下方的距离

        \caption{Left: Lewis Game. Right: Multi-token Discrete Mechanism. The communication vector is initially divided into multiple discretization tokens. Each token goes through separate discretization, where it is quantized to the nearest neighbor within a shared collection of latent codebook vectors. Subsequently, the discretization tokens are concatenated back together to form a vector with the same shape as the original one.}
        \label{fig1}
        % \vspace{-0.2cm}
\end{figure}

Intelligent agent communication, positioned at the intersection of AI and linguistics, explores the development of a shared language among agents. The Lewis Game\cite{bib1} serves as a key example of collaborative tasks in this area, where a speaker and listener work together to identify a specific object from a set of alternatives (Figure \ref{fig1}(left)). This field has seen extensive research into the origins and evolution of language through AI, with studies investigating various aspects of emergent communication. \cite{bib5} focused on the compositionality and generalization capabilities of language agents and the creation of efficient color naming systems\cite{bib7}. \cite{bib8} proposed blending multi-agent communication with data-driven natural language learning to facilitate machine-human interaction. \cite{bib9} explored the interplay between supervised learning and self-play in developing communication protocols for emergent communication, while deep reinforcement learning has been applied to tackle the challenges of multi-agent communication\cite{bib10,bib11}, especially as the number of agents increases, leading to potential redundancy and inefficiency in communication. Recent research\cite{bib12,bib13,bib14,bib15} has focused on optimizing communication strategies, expanding our understanding of language evolution by addressing both capacity and dataset complexity challenges. Vector Quantization (VQ), in line with Shannon's rate-distortion theory\cite{bib16}, suggests that vector encoding can outperform scalar encoding by effectively handling dependencies in source symbols. Recent advancements in reparameterization, particularly for VAEs managing discrete variables\cite{bib17,bib18}, have enhanced model effectiveness. The VQ-VAE\cite{bib19} model overcomes non-differentiability issues by employing the identity function for efficient gradient transmission, proving to be more robust and generalizable than continuous counterparts for complex learning models\cite{bib19,bib20}. Moreover, the application of discretization in multi-agent reinforcement learning tackles communication challenges within modular reasoning architectures, facilitating efficient interactions across modules\cite{bib21}.

This study builds on previous research by advancing the vector quantization technique in VQ models. Our hypotheses suggest that when intelligent agents have diverse personal experiences, communicating via discrete messages, especially through sentences composed of multiple discrete tokens, is more effective than using continuous messages. Our experiments in multi-agent machine learning have empirically demonstrated that communication through multi-token discrete sentences significantly enhances communication efficiency among agents with diverse experiences. However, when employing the VAE model to simulate continuous language communication between agents, rather than the AE model, its effectiveness surpasses that of multi-token discrete language communication. 

The structure of our paper is organized as follows: Section \ref{sec1} provides a background on agents'  communication and the objectives of our research. Section \ref{sec2} introduces related work in the field. Section \ref{sec3} details the experimental settings and methods employed in our study. In Section \ref{sec4}, we present our experimental results and draw meaningful conclusions. Finally, Section \ref{sec5} concludes our research with discussions and prospects for future work.
\vspace{-0.2cm}
\section{Related Work }\label{sec2}
In recent years, several methods have been devised to improve communication within machine learning models, notably through attention mechanisms\cite{bib21,bib22,bib23} and the Transform method\cite{bib24,bib25}. Furthermore, collective memory and shared parameters have enhanced multi-agent communication\cite{bib26}.
The Reinforced Inter-Agent Learning (RIAL) model stands as a prominent framework for discrete communication among intelligent agents, supporting the use of discrete symbols to enable interactions reminiscent of human social behaviors\cite{bib11}. The work of \cite{bib27} delves into the crucial role of computer simulations in evolutionary linguistics, illustrating how intelligent agent models can foster the development of compositional languages for numerical concepts through communication. Research into LSTM\cite{bib28} language models by \cite{bib29} has illuminated the manner in which hidden states encapsulate numerical values and syntactic structures, spurring further exploration into linguistic patterns. Subsequent studies\cite{bib30,bib31,bib32} have expanded our comprehension of the dynamics of multi-agent communication and the genesis of language among neural network-based agents.

Building on this foundation, our research leverages the Vector-Quantized Variational Autoencoder (VQ-VAE) model\cite{bib19} and adopts cross-training and cross-validation techniques to scrutinize communication patterns between agents. Our findings reveal that in environments where agents employ diverse language systems, discrete forms of language are more efficacious than continuous ones. We also delve into how the variation in token numbers within codebooks affects the efficiency of discrete communication, a subject that receives in-depth treatment in Section \ref{sec4}.
\section{Theoretical Basis and Experimental Method}\label{sec3}
% In the series of autoencoder models\cite{ackley1985learning}, we designate the encoder component of the model as the speaker, while the decoder component is referred to as the listener. In this framework, when given an input, the speaker transforms it into a sequence of information and transmits it to the listener. The listener then reconstructs the information based on the received messages. The speaker and listener collaborate in the learning process, aiming to minimize the reconstruction loss. During this process, the continuous data output by the encoder is directly input into the decoder, which is the process of agents using continuous language for communication.
In the series of autoencoder models\cite{bib33}, we designate the encoder component of the model as the speaker, while the decoder component is referred to as the listener. The communication between the speaker and the listener within the same model is considered internal communication, while the communication between the speaker or listener and other model components is referred to as agents' communication. We define $e\left ( \cdot  \right )$  as the encoder function, $d\left ( \cdot  \right )$  as the decoder function, and $h\left ( \cdot  \right )$ as the quantization layer function of VQ-VAE. The codebook space of the quantization layer is denoted as $C=\left [   c_{1}, c_{2},...c_{n}  \right ]$.
\subsection{Discrete and Continuous Communication}\label{subsec1}
For the discrete communication model, we use VQ-VAE. To train a pair of agents, let's assume the input is $x$. The information is passed through the speaker as $z_{e}(x)= e(x,\theta )$, representing the encoded representation of $x$. Then, the information undergoes discrete quantization using a codebook, resulting in $Z=h(z_{e}(x),\varphi)$. Finally, the speaker reconstructs the original information from the received codebook indices $x^{'}=d(Z,\phi)$, where the model parameters $\theta,\varphi,\phi$  are continuously updated by minimizing the reconstruction loss and codebook loss. The complete loss function for this process is as shown in Equation \ref{eq1}:
\begin{equation}
\label{eq1}
\mathbb{L}_{VQ}=||x-x^{'} ||_{2} +||sg[z_{e}(x)]-e_{k} ||_{2}^{2}+\beta||z_{e}(x)-sg[e_{k}]||_{2}^{2}
\end{equation}
Where the last two terms represent the quantization loss in the VQ-VAE model, in the subsequent algorithm, we use $L_{quantify}$ to represent these two items. 
In the experiments involving the AE model, the overall loss can be expressed as:$z=e(x,\theta )$,$x^{'}=d(z,\phi )$. As shown in Equation \ref{eq2}, the overall loss is equivalent to the reconstruction loss.
\begin{equation}
\label{eq2}
\mathbb{L}_{AE}=||x-x^{'} ||_{2}=||x-d(e(x,\theta ),\phi )||_{2} 
\end{equation}
During this process, the continuous data output by the encoder is directly input into the decoder, which is the process of agents using continuous language for communication. And the involvement of the codebook quantization layer mentioned above refers to the process of discrete communication.

Throughout the entire experiment, the experimental data based on the Autoencoder (AE) serves as a baseline, which aims to verify that under the same experimental settings, the use of continuous communication is less effective than discrete communication between unfamiliar agents.
\subsection{Multi-token Discretization}\label{subsec2}
\cite{bib34} proposed a human-like discrete information generation method that enables discrete message communication to have the effect of continuous message communication. Based on the foundation of discretization, we propose a multi-token discretization approach. The VQ-VAE model builds upon the AE and introduces a latent space codebook between the encoder and decoder. In our research, multi-token discretization is applied before the data enters the codebook layer. It involves dividing the output of the encoder into multiple segments of equal size but containing different data. Let's assume our latent codebook size is $e\in R^{L\times M} $. Initially, the output $z_{e}(x)$ is divided into $N$ segments $s_{1}, s_{2},s_{3},...s_{N}$ with $z_{e}(x)=CONCAT(s_{1}, s_{2},s_{3},...s_{N})$, where each segment $s_{i} \in \mathbb{R}^{\frac{M}{N} } $ with $\frac{M}{N}\in N^{+  } $. Next, each of these segments is discretized sequentially: $e_{o_{i} }=h(s_{i} ) $, where $o_{i}=\underset{j}{argmin} \left \| s_{i}-c_{j} \right \| _{2}$ . After the discretization process, the $N$ segments of data are then integrated back together in the order of their original splitting: $Z=CONCAT(e_{o_{1} }, e_{o_{2} }, ...e_{o_{N} })$. 
Throughout the entire process, the discretized multi-token data always shares the same codebook. The schematic diagram of the multi-token discretization is illustrated in Figure \ref{fig1}(Right). Since we divide the data into $N$ segments, the total loss function for model training is defined as shown in Equation \ref{eq3}.
\begin{equation}
\label{eq3}
\mathbb{L}= \mathbb{L}_{task}  +\frac{1}{N}\left \{\sum_{i=1}^{N}||sg[s_{i} ]-e_{o_{i} } ||_{2}^{2}+\beta\sum_{i=1}^{N}||s_{i} -sg[e_{o_{i} }]||_{2}^{2}\right \}  
\end{equation}
Where $L_{task}$ represents the specific task loss, which can be the aforementioned reconstruction loss, classification loss, or any other relevant loss function.
\begin{figure}[htbp]
	% \centering
        % \vspace{-0.55cm}  %调整图片与上文的垂直距离
	\begin{minipage}{0.5\linewidth}
		\centering
		\includegraphics[width=1.0\linewidth]{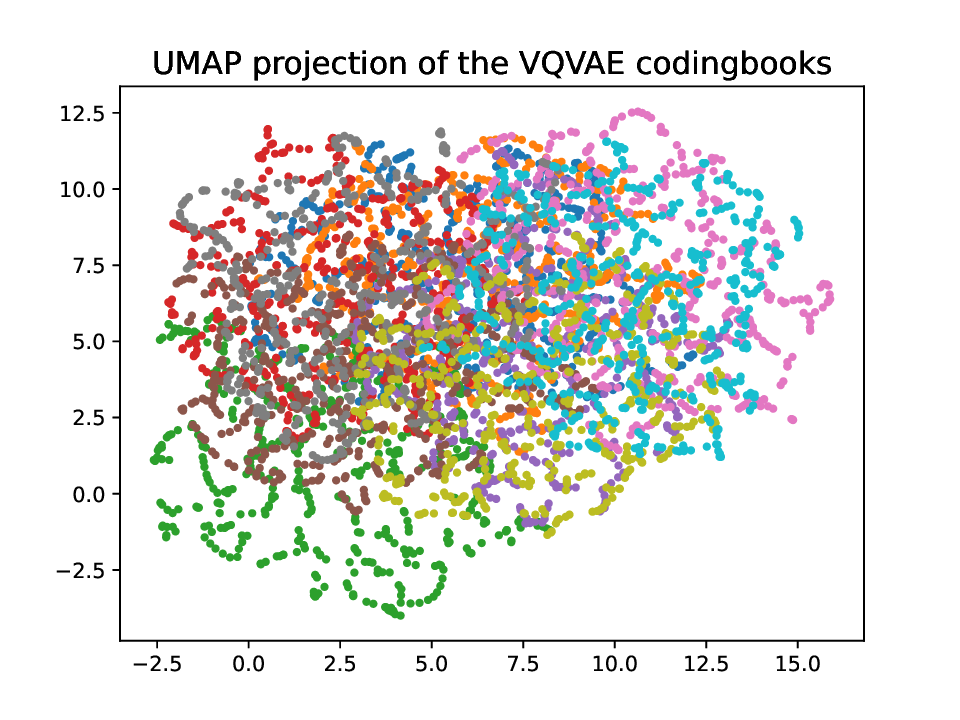}
		% \caption{图1}
		% \label{图1}%文中引用该图片代号
	\end{minipage}
        \begin{minipage}{0.4\linewidth}
		\centering
		\includegraphics[width=1.2\linewidth]{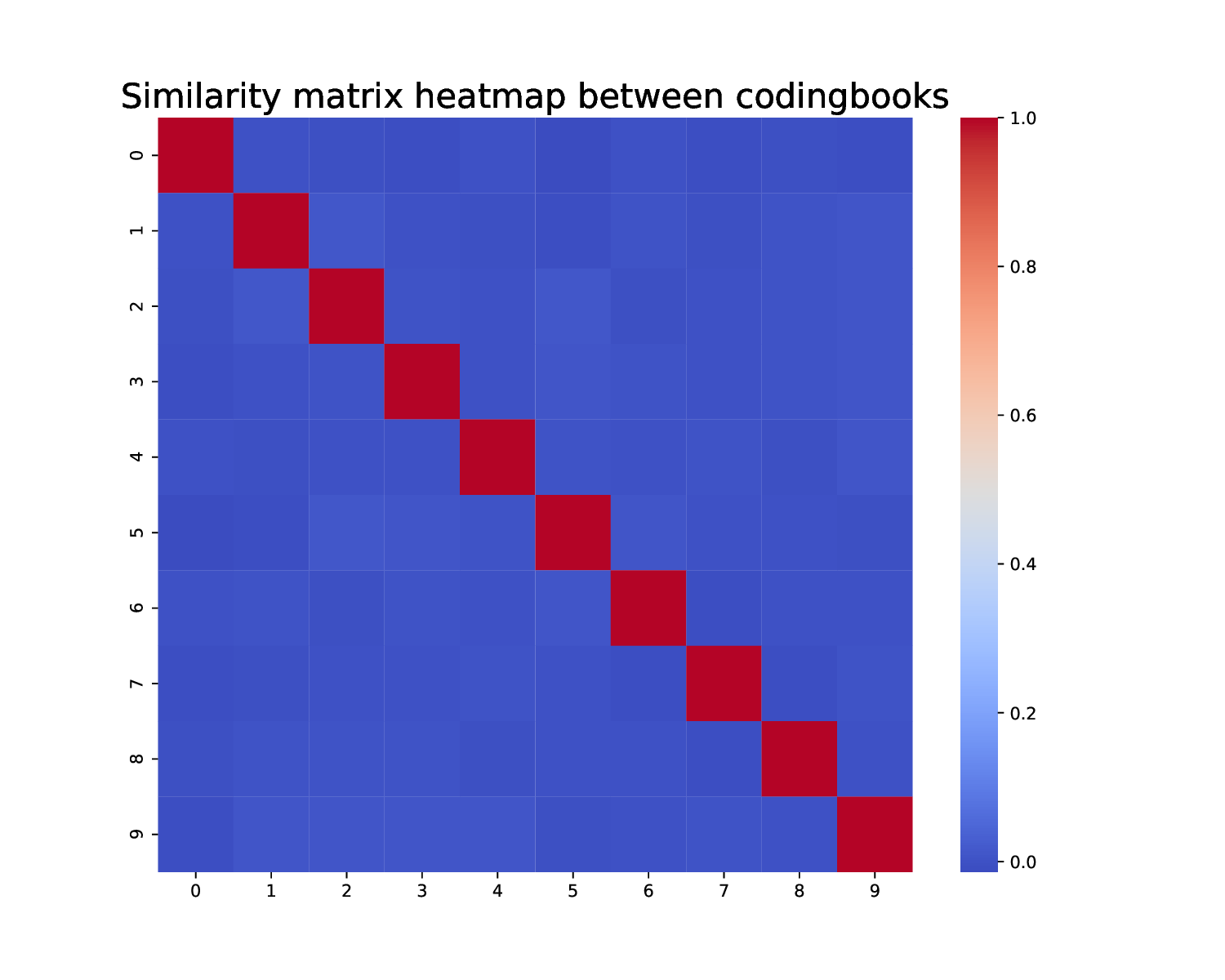}
		% \caption{图2}
		% \label{图2}%文中引用该图片代号
	\end{minipage}
        % \centering
	% \setlength{\abovecaptionskip}{0.cm} %调整标题上方的距离
        \setlength{\abovecaptionskip}{0cm} %调整标题下方的距离
        \caption{Ten agents' understanding of the same language. Left: Feature distribution of latent codebooks for different agents. Right: Similarity of different latent codebooks. The understanding of this language is different for each agent.}
        \label{fig2}
        % \vspace{-0.55cm}
\end{figure}

\subsection{Learning and Validation of Communication for Agents}\label{subsec3}
Attempts have been made to explore cross-training\cite{bib27,bib35} in the context of multi-agent learning. In this approach, during the simultaneous training of multiple agents, after each iteration, a random combination is selected, pairing one agent's speaker with another agent's listener for the next round of iterative learning. The reason behind this approach is that when multiple agents learn the same language, their understanding of the language may not be entirely identical. Figure \ref{fig2} demonstrate the feature distributions of the latent codebook spaces for $10$ agents trained simultaneously on the same MNIST dataset. Each color represents a internal communication protocol, that is, the feature distribution in the codebook. It can be observed that there are differences in semantic understanding among agents. Hence, cross-training becomes necessary because it allows different agents to have the most similar understanding of the same language. Algorithm \ref{algo2} in the Appendix \ref{appendix} implements the aforementioned process.
% \vspace{-0.2cm}
\begin{algorithm}
\caption{Individual Training and communication} \label{algo1}
\begin{algorithmic}[1]

\State {Using the processed dataset: $train_{-}set =(train_{-}set_{1},train_{-}set_{2},...train_{-}set_{m})$}

\State {Train $j$ agents simultaneously}

\State {Initialize encoders $E = \{e_0\left ( \cdot  \right ), ..., e_{m}\left ( \cdot  \right )\}$.}

\State {Initialize quantization layers $H = \{h_0\left ( \cdot  \right ), ..., h_{m}\left ( \cdot  \right )\}$} 

\State {Initialize decoders $D = \{d_0\left ( \cdot  \right ), ..., d_{m}\left ( \cdot  \right )\}$}

\For {each iteration $i$}

    \For {each agent $j$}

\State {Sample input data $x_{j}$ from $train_{-}set_{j}$}
   
    \State {$L_{j} \Leftarrow MSE(x_{j}, d_{j} (h_{j}(e_{j}(x_{j} ) ) )$}

    \State {optimize $e_{j}\left ( \cdot  \right )$,$h_{j}\left ( \cdot  \right )$ and $d_{j}\left ( \cdot  \right )$ with respect to $L_{j}$}
    
    \EndFor
\EndFor

\State{Validation:}
  
%   \Indm
\For {each agent $j$}
   
    \State {Sample validate data $x$ from $val_{-}set$}
    \State {$loss_{j} \Leftarrow MSE(x,d_{k} (h_{j}(e_{j}(x) ) ) ),(k=1,2...m,k\neq j)$}
    \State {output $loss_{j}$}

\EndFor
\State {$Communication\frac{}{} loss = \sum loss_{j} $}

\end{algorithmic}
\end{algorithm}

\begin{figure}[htbp]
% \vspace{-0.35cm}
\centering
\includegraphics[width=1.1\linewidth]{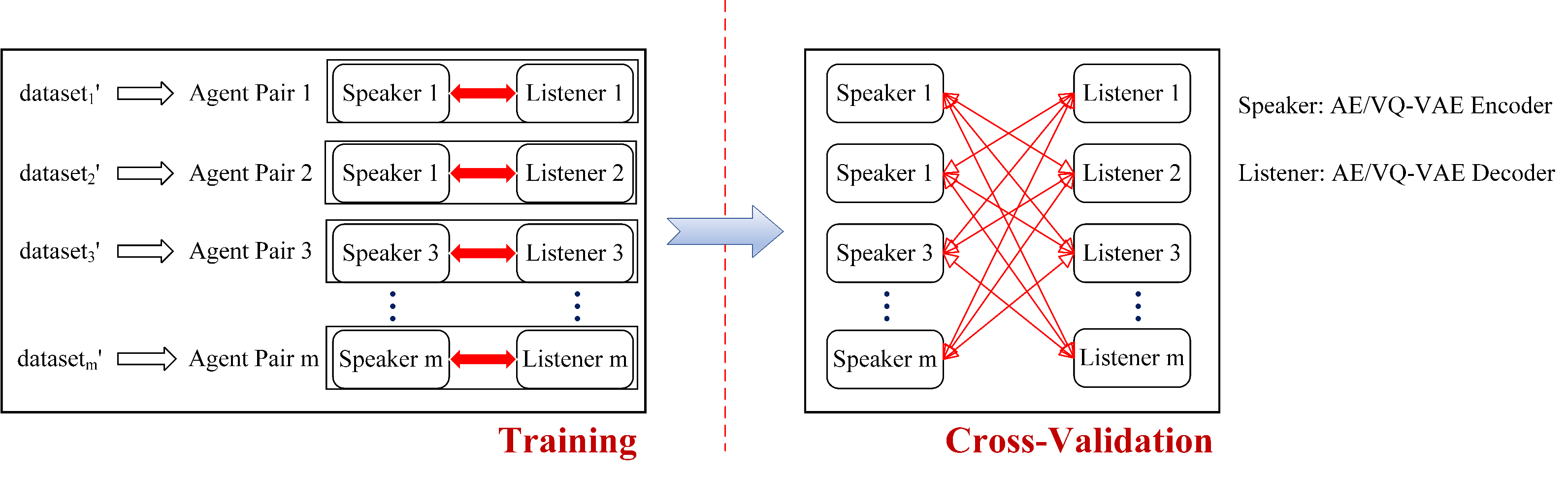}
\setlength{\abovecaptionskip}{-0.50cm}
\caption{Training and validation of agents. Each agent has its own dataset during training. Upon completion of learning, one agent interacts with the other agent . This is our core methodology, where in this validation scenario, the advantage of discrete language in communication between agents is determined based on the reconstruction losses of information.}
\label{fig3}
% \setlength{\abovecaptionskip}{-0.35cm}
% \vspace{-0.50cm}
\end{figure}

% \vspace{-0.2cm}
In terms of experimental validation, in addition to using trained agents for verification, we conducted another form of validation by manipulating the dataset. Assuming there are $m$ types of data in the dataset, we merged the training set and validation set into a single dataset. The merged dataset was then divided into $m$ classes based on their labels $Dataset=(dataset_{1},dataset_{2},...dataset_{m})$. After the division, a portion of images was uniformly sampled from each class to form the validation set, denoted as $val_{- } set=(sample_{1}^{1} ,sample_{2}^{1},...sample_{m}^{1})$. To meet the experimental overlap requirements, from the remaining training set $P_{train}$, a certain number of images were extracted from each class according to the desired experimental overlap rate across the classes $Overlapset=(sample_{1}^{2} ,sample_{2}^{2},...sample_{m}^{2})$. There are an equal number of images in each class, and $P_{train}$ is the number of images left after the first extraction. For $p_{j}\in \left \{ 0.05,0.1,0.2,...0.9 \right \}$, the calculation of the number of images sampled in the second extraction is given by Equation \ref{eq4}. 
\begin{equation}
\label{eq4}
\frac{m\ast sample_{i}^{2}  }{P_{train}+(m-1)*sample_{i}^{2} } =p_{j} ,(i=1,2...,m)
\end{equation}
$Overlapset$ were then merged with the respective training sets $dataset_{i}^{'}$, where $dataset_{i}^{'} = dataset_{i}-sample_{i}^{1}-sample_{i}^{2}$, ensuring that each class in the training set contained images from the remaining $m-1$ classes. The processed training set is $train_{-}set =(train_{-}set_{1},train_{-}set_{2},...train_{-}set_{m})$. This process resulted in a training set where each category served as a separate training set for single agent to learn from. The experiments conducted on the split dataset follow Algorithm \ref{algo1}, which forms the core of our research paper. Similar to Algorithm \ref{algo1}, the experimental methodology of our core content is illustrated in Figure \ref{fig3}.

The aforementioned are the two experimental procedures we used to explore the communication patterns of multiple agents. The main model involved in the procedures is the VQ-VAE model. However, when we incorporate the AE model in our experiments, we simply remove quantization layers $H$ from the procedure, and the data outputted by the encoder is directly decoded by the decoder.
\section{Experiments}\label{sec4}
\begin{figure}[htbp]
	\centering
        % \vspace{-0.35cm}  %调整图片与上文的垂直距离 
	\begin{minipage}{0.49\linewidth}
		\centering
		\includegraphics[width=0.8\linewidth]{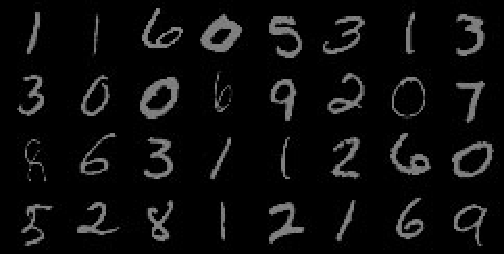}
		% \caption{图1}
		\label{图1}%文中引用该图片代号
	\end{minipage}
        \begin{minipage}{0.49\linewidth}
		\centering
		\includegraphics[width=0.8\linewidth]{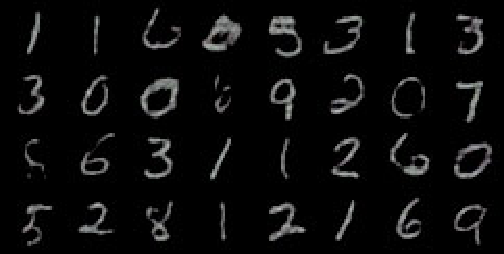}
		% \caption{图2}
		\label{图2}%文中引用该图片代号
	\end{minipage}
        % \centering
        \caption{Example images for the reconstruction task. Left: Original image. Right: Reconstructed image.}
        % \vspace{-0.2cm}  %调整图片与上文的垂直距离
        \label{fig4}
\end{figure}

 In our work, we employed four datasets: MNIST, CIFAR10, CelebA and Diabetic Retinopathy dataset. The image resolution for all four datasets was separately set to $28\times 28,32\times 32,64\times 64$ and $64\times 64$. The batch size for the first two datasets during training is set to $256$, while the batch size for the latter two datasets is set to $64$, and we utilized the Adam optimizer with a learning rate of $0.001$. The commitment cost for the model's discrete layer was set to $0.25$, with a decay rate of $0.99$. The specific codebook size for the discrete layer varied depending on the dataset. In the experiments, we evaluate the effectiveness of agents' communication by measuring the error between the original images and the reconstructed images. An example of the two types of images can be seen in Figure \ref{fig4}.
\subsection{Multi-token Discretization for Improved Agents' Communication}\label{subsec1}
\begin{figure}[htbp]
    \vspace{-0.25cm}  %调整图片与上文的垂直距离
    \centering
    \includegraphics[width=0.32\linewidth]{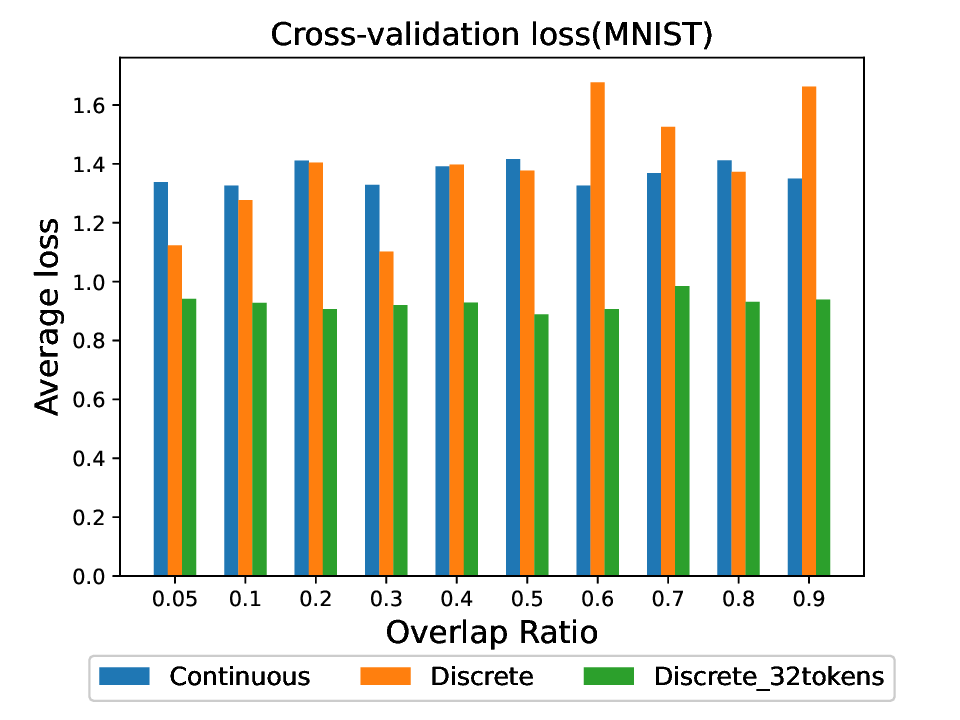}
    \hfill
    \includegraphics[width=0.32\linewidth]{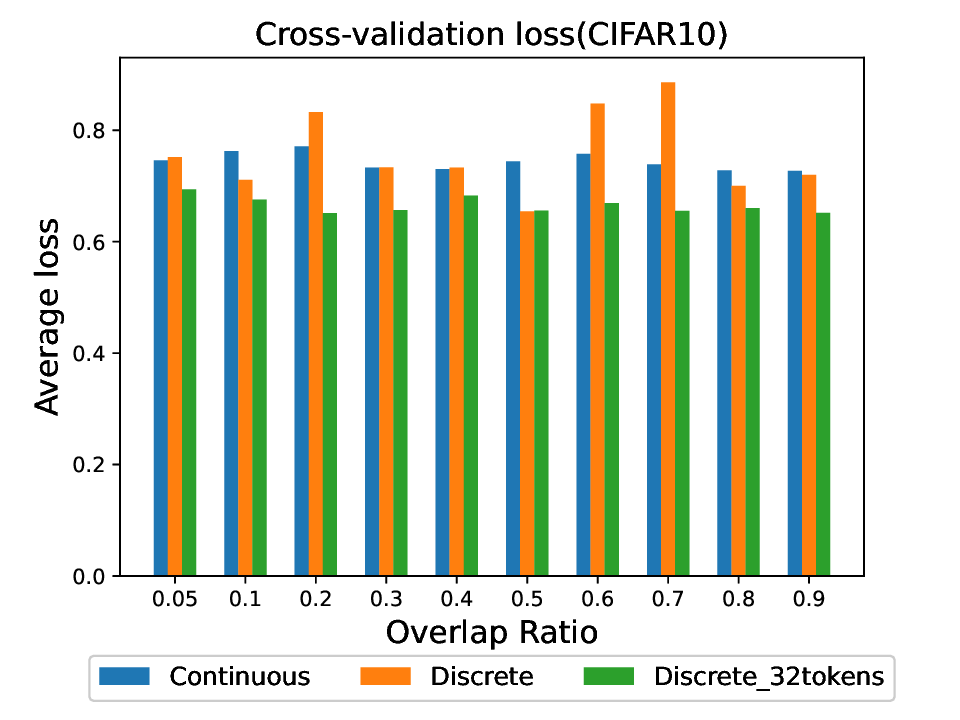}
    \hfill
    \includegraphics[width=0.32\linewidth]{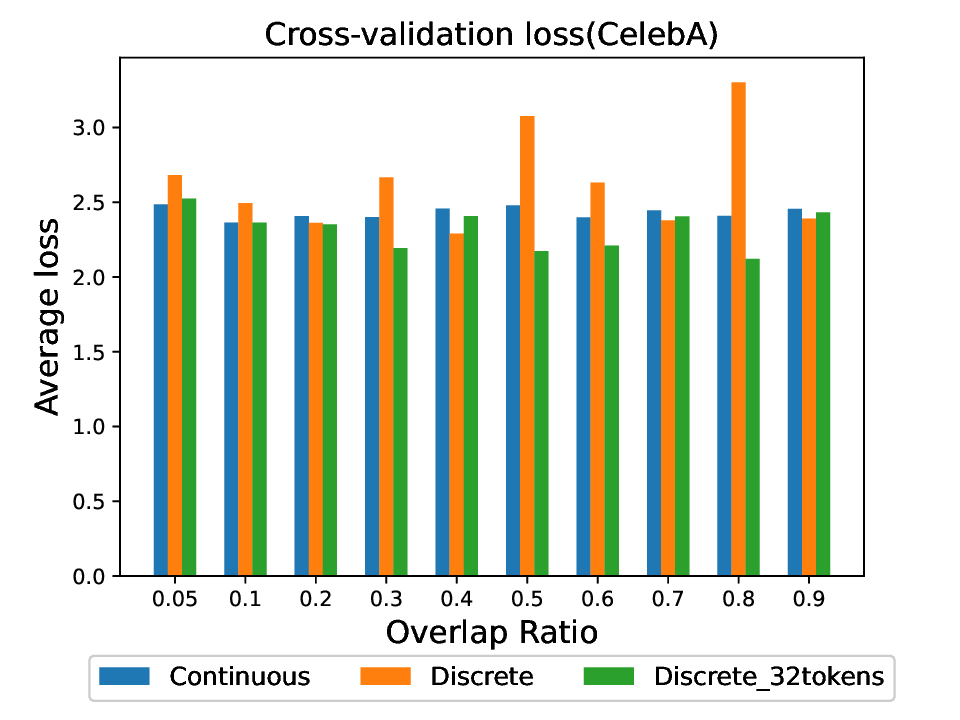}
    \caption{Communication loss on three types of models. The x-axis represents different overlap ratios, and the y-axis represents the communication loss between agents.}
    % \vspace{-0.25cm}  %调整图片与上文的垂直距离
    \label{fig5}
\end{figure}

\begin{figure}[htbp]
    \vspace{-0.25cm}  %调整图片与上文的垂直距离
    \centering
    \includegraphics[width=0.32\linewidth]{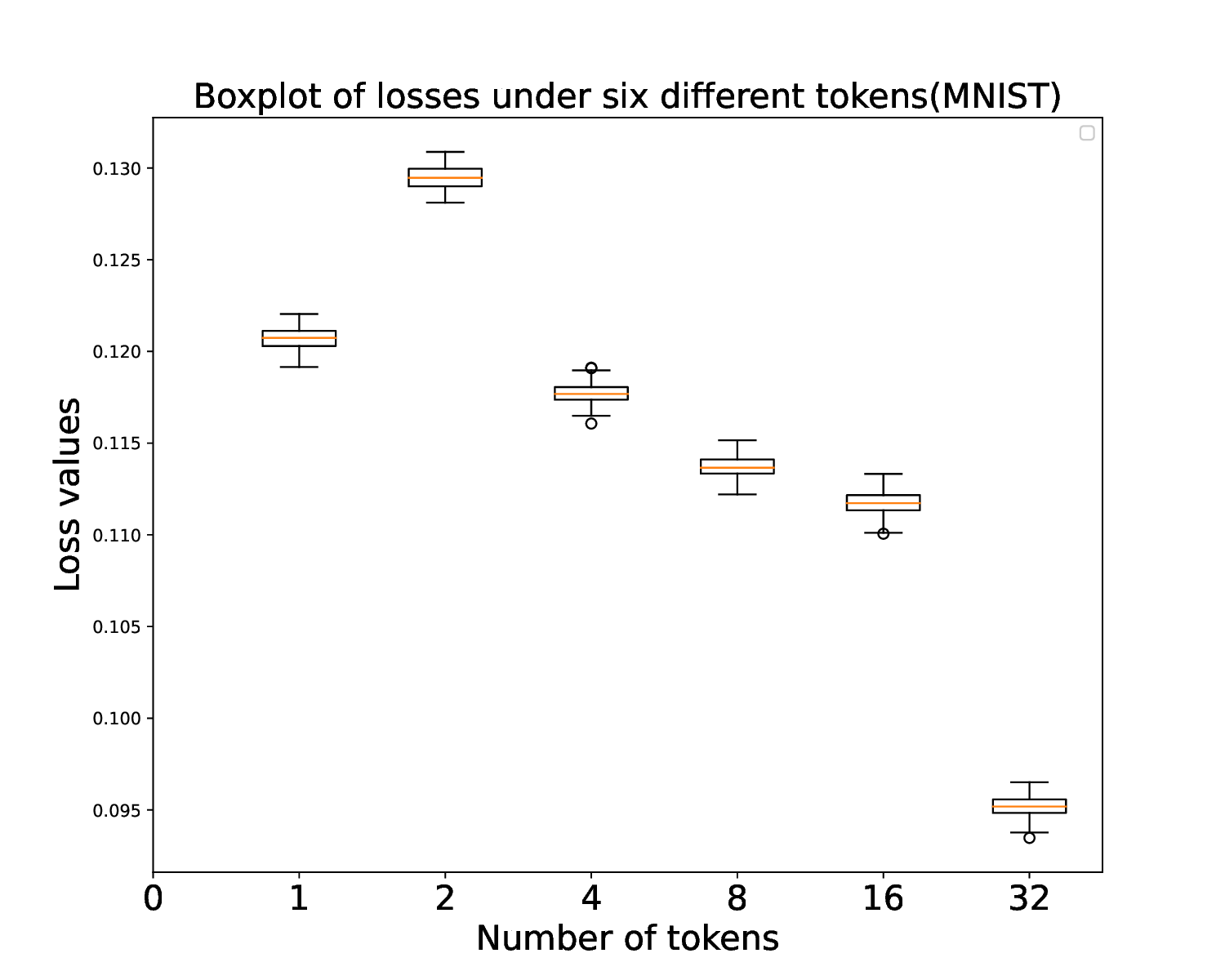}
    \hfill
    \includegraphics[width=0.32\linewidth]{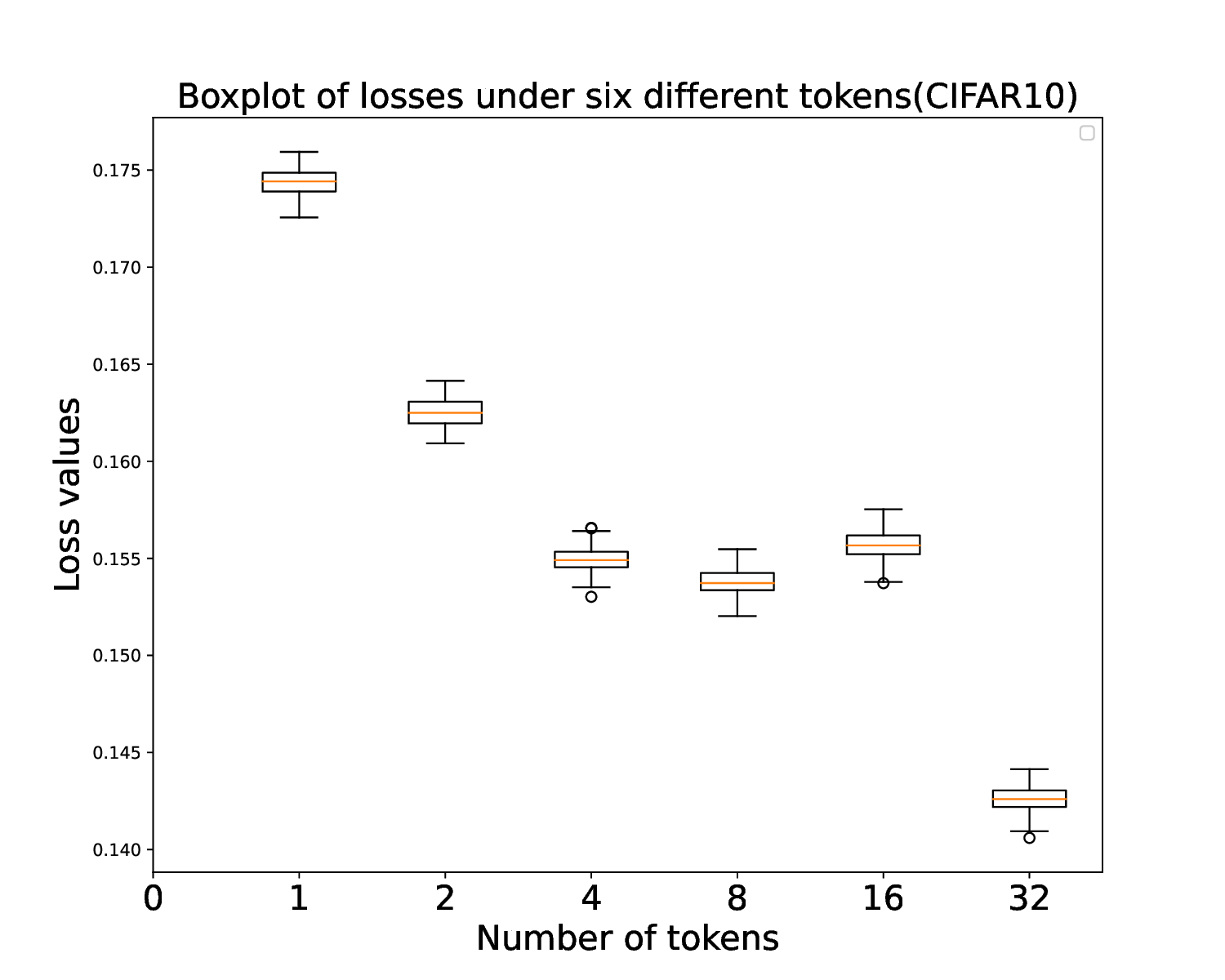}
    \hfill
    \includegraphics[width=0.32\linewidth]{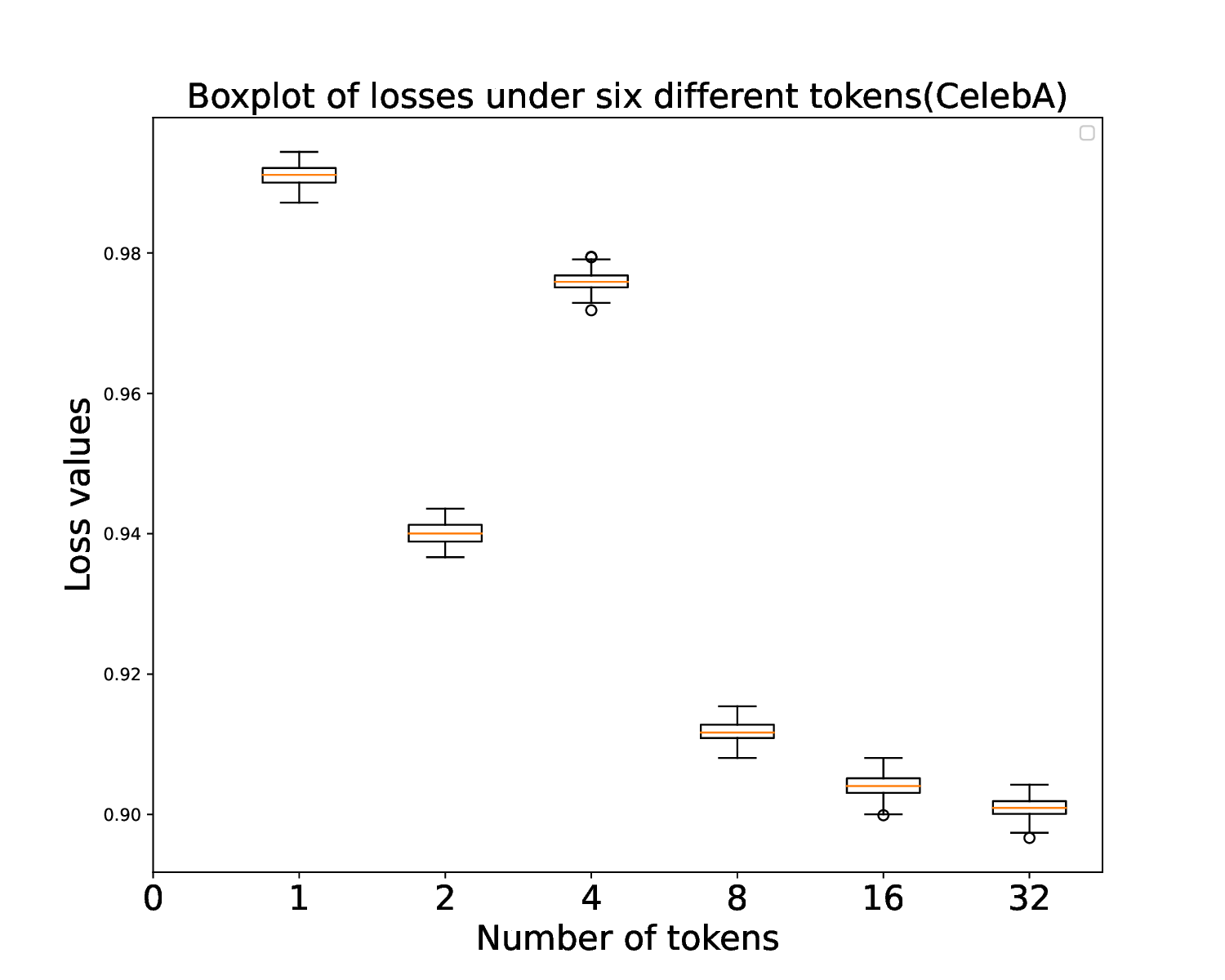}
    \setlength{\abovecaptionskip}{-0.1cm} %调整标题下方的距离
    \caption{The communication loss of multiple agents with the same data overlap ratio under multi-token discretization. The increase in the number of tokens can reduce the loss from communications. }
    % \vspace{-0.25cm}  %调整图片与上文的垂直距离
    \label{fig6}
\end{figure}

According to the method shown in Figure \ref{fig3} and Algorithm \ref{algo1}, we repeated the experiments with different overlap ratios using the multi-head discretized VQ-VAE model with the best performance and the AE model. For the MNIST and CelebA datasets, the original VQ-VAE model had  latent space size $C_{m} \in \mathbb{R} ^{512\times 64}$ and $C_{m} \in \mathbb{R} ^{512\times 128}$, while for the CIFAR10 dataset, it was $C_{m} \in \mathbb{R} ^{1024\times 256}$. We conducted the above overlap experiments with $32$ tokens, and the experimental results are shown in Figure \ref{fig5}. Under three different datasets, the average loss incurred by using multiple discrete tokens for communication is 32.1\%, 10.6\%, and 3.7\% lower than that incurred by using continuous semantics for communication, respectively. Our experiments indicate that when one agent interacts with another unfamiliar agent, the discrete semantic learning method using multiple tokens has certain advantages over continuous semantic learning. 
Figure \ref{fig6} explain why we chose to conduct our experiments with a 32-token VQ-VAE model and also illustrate the advantages of our proposed multi-token discrete mechanism compared to a single-token approach. It shows the results of training $m$ agents simultaneously according to Algorithm \ref{algo1}, where each boxplot in the figure represents the stable loss from communications between the $m$ agents. The general pattern is that as the number of discrete tokens increases, the communication loss decreases.

In all of the above experiments, the number of agents $m$ for the three datasets respectively are ($m=10(MNIST,CIFAR10),8(CelebA)$). Our experiments have demonstrated two theories. First, discrete communication with multiple tokens are more effective than continuous ones when agents have diverse personal experiences. Second, communications using multiple discrete tokens are more advantageous than those using a single token. 

\subsection{Theoretical Validation and Practical Application}\label{subsec2}
\begin{figure}[htbp]
	% \centering
        % \vspace{-0.45cm}
	\begin{minipage}{0.49\linewidth}
		\centering
		\includegraphics[width=0.8\linewidth]{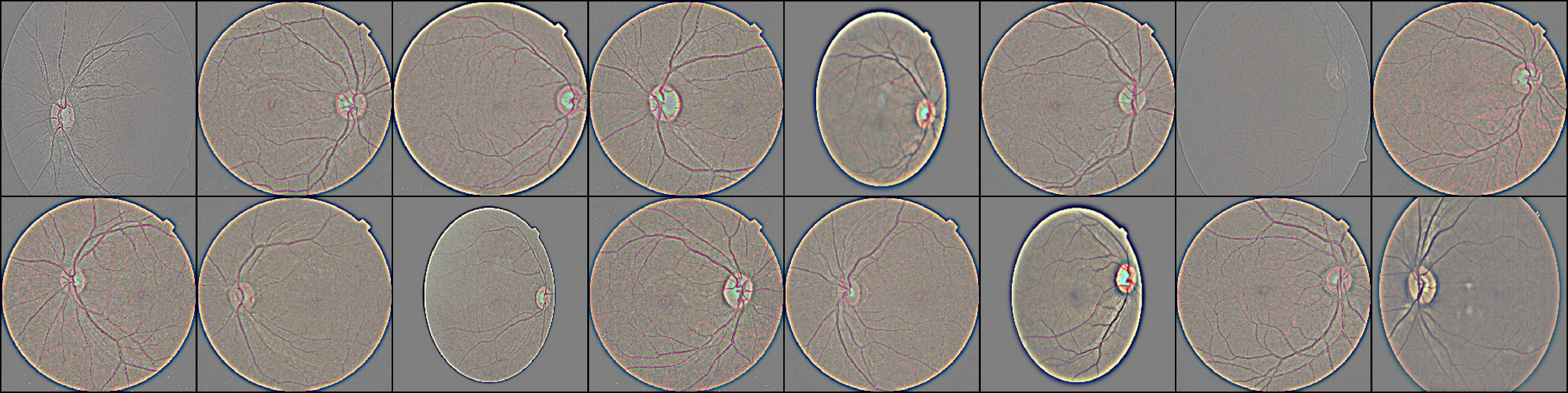}
		% \caption{图1}
		\label{图1}%文中引用该图片代号
	\end{minipage}
        \begin{minipage}{0.49\linewidth}
		\centering
		\includegraphics[width=0.8\linewidth]{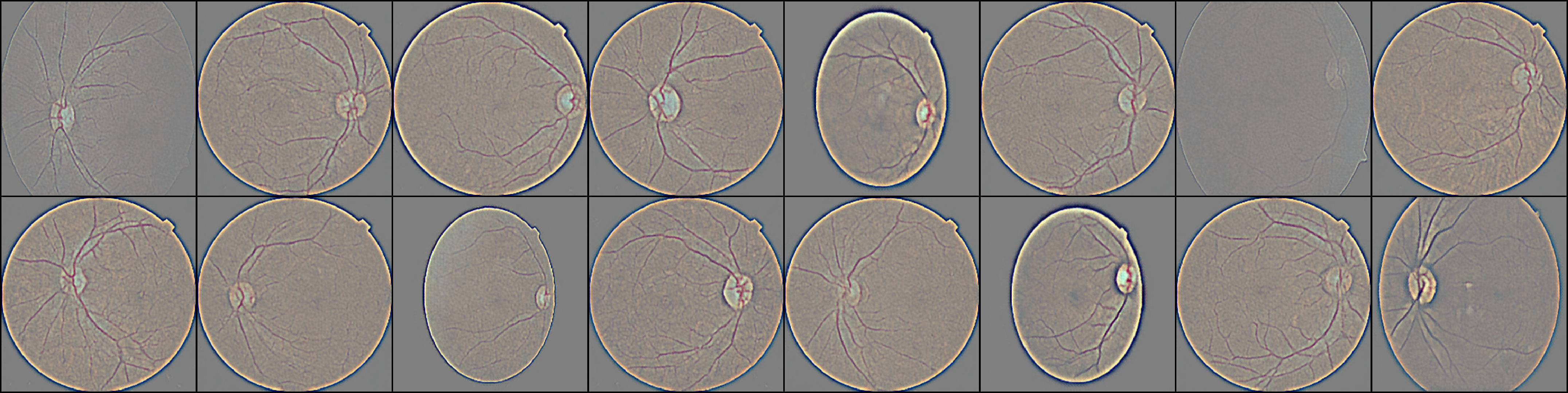}
		\label{图2}%文中引用该图片代号
	\end{minipage}
        % \centering
        \setlength{\abovecaptionskip}{0.1cm} %调整标题下方的距离
        \caption{Sample images of medical dataset. Left: Original image. Right: Reconstructed image.}
        % \vspace{-0.4cm}  %调整图片与上文的垂直距离
        \label{fig7}
\end{figure}

Based on the open-source datasets, we conducted the same experiments as in section \ref{subsec1} with Diabetic Retinopathy dataset(Figure \ref{fig7}) to further validate our theory. When experimenting with the VQ-VAE model on this dataset, codebook size $e_{m} \in \mathbb{R}^{512\times 128}$. The dataset is divided into $5$ categories based on symptom types, with varying numbers of images in each category. We performed $5$ sets of experiments, each involving communications between agents. 
% First, to demonstrate the feasibility of our proposed multi-token discretization, we conducted experiments following Algorithm \ref{algo1}. The experimental results are shown in Figure \ref{fig8}(Right), which once again validate the effectiveness of our theory.

\begin{figure}[htbp]
\vspace{-0.5cm}  %调整图片与上文的垂直距离
	\begin{minipage}{0.49\linewidth}
		\centering
		\includegraphics[width=0.8\linewidth]{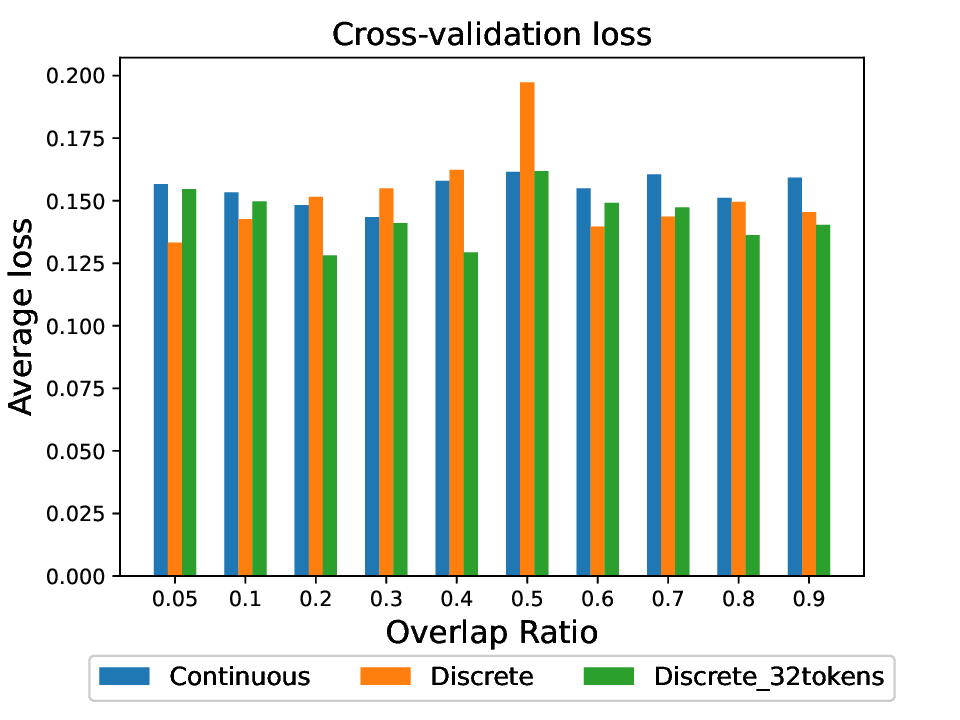}
		% \caption{图1}
		\label{图1}%文中引用该图片代号
	\end{minipage}
        \begin{minipage}{0.49\linewidth}
		\centering
		\includegraphics[width=0.8\linewidth]{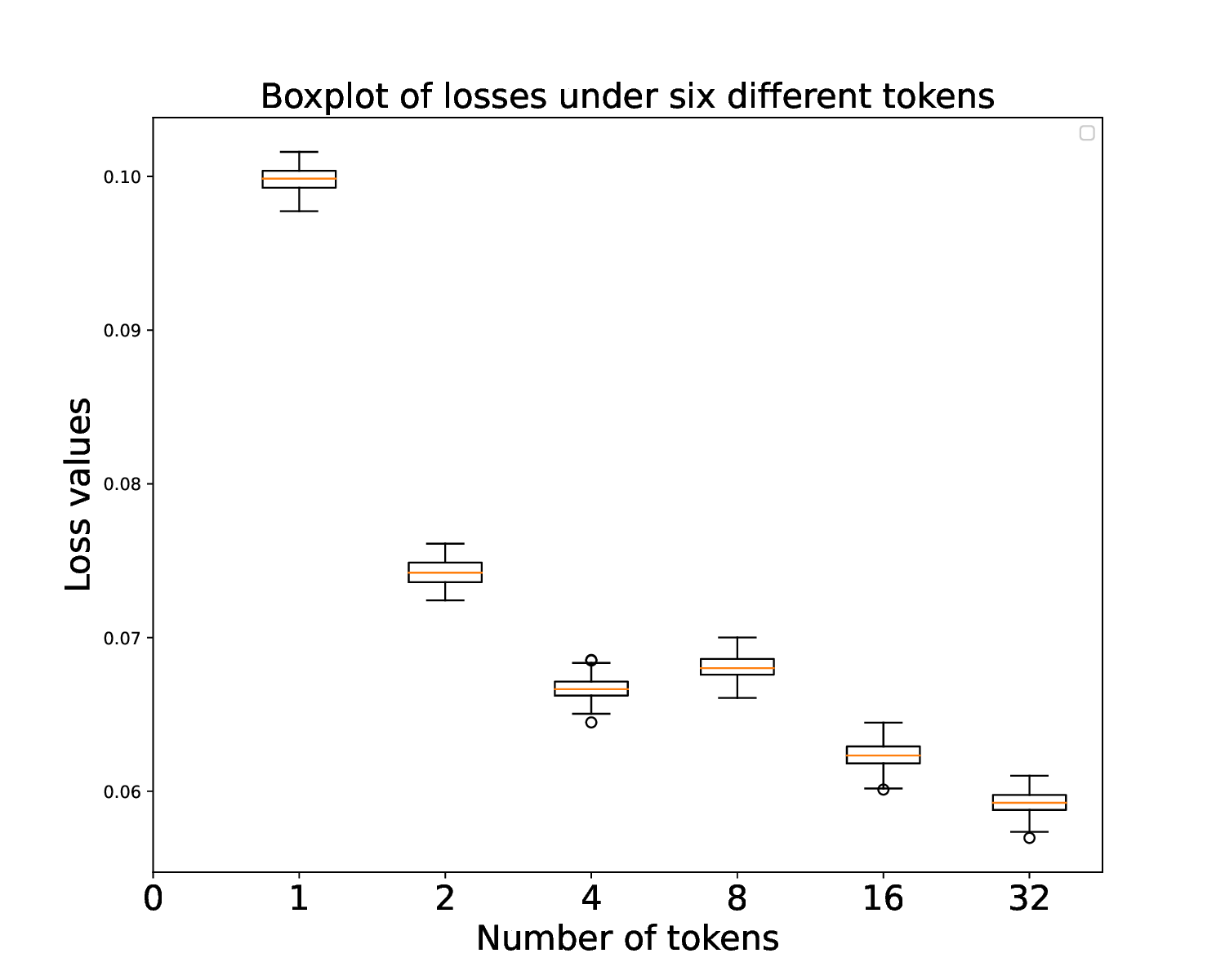}
		% \caption{图2}
		\label{图2}%文中引用该图片代号
	\end{minipage}
        \setlength{\abovecaptionskip}{0.1cm} %调整标题下方的距离
    \caption{Experimental results under Diabetic Retinopathy dataset. Left: Communication losses on three types of models(AE, VQ-VAE, VQ-VAE-32token); Right: Communication losses of multiple agents under multi-token discretization.}
    % \vspace{-0.2cm}  %调整图片与上文的垂直距离
    \label{fig8}
\end{figure}

The dataset, which originally consisted of only $2750$ images, has been expanded to $5000$ images by applying data augmentation techniques. Each class now contains $1000$ augmented images. First, we processed the dataset according to the data preprocessing steps outlined in Algorithm \ref{algo1}, and completed the communication-validation experiments. The communication loss under multi-token discretization is shown in Figure \ref{fig8}(Right).

Then, we conducted experiments on the core theoretical aspects based on this dataset. The results, shown in Figure \ref{fig8}(Left), indicate that when the number of discrete tokens reaches $32$, the overall discrete interactive communication outperforms continuous interactive communication, the former's average loss is 7.1\% lower than that of the latter. The experimental results on the new dataset provided strong evidence to support our conclusions. Communication between agents who are unfamiliar with each other using multi-token discretized information variables is better than using continuous variables.

\subsection{Research on codebook aspects}\label{subsec3}
\begin{figure}[htbp]
	% \centering
        \vspace{-0.25cm}  %调整图片与上文的垂直距离
	\begin{minipage}{0.48\linewidth}
		\centering
		\includegraphics[width=0.8\linewidth]{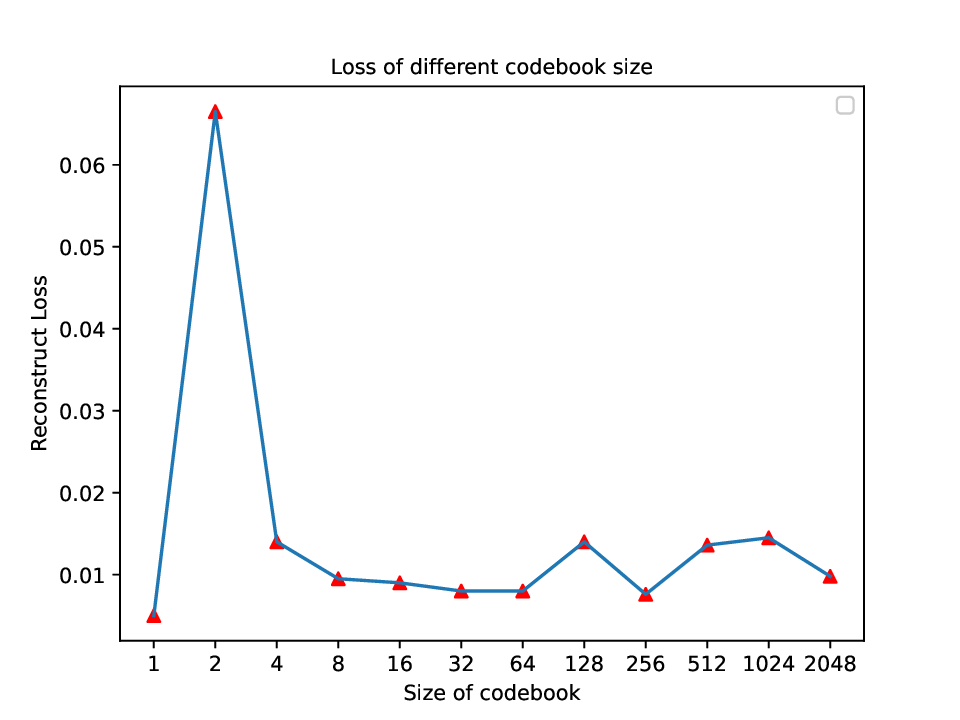}
		% \caption{图1}
		\label{图1}%文中引用该图片代号
	\end{minipage}
        \begin{minipage}{0.48\linewidth}
		\centering
		\includegraphics[width=0.8\linewidth]{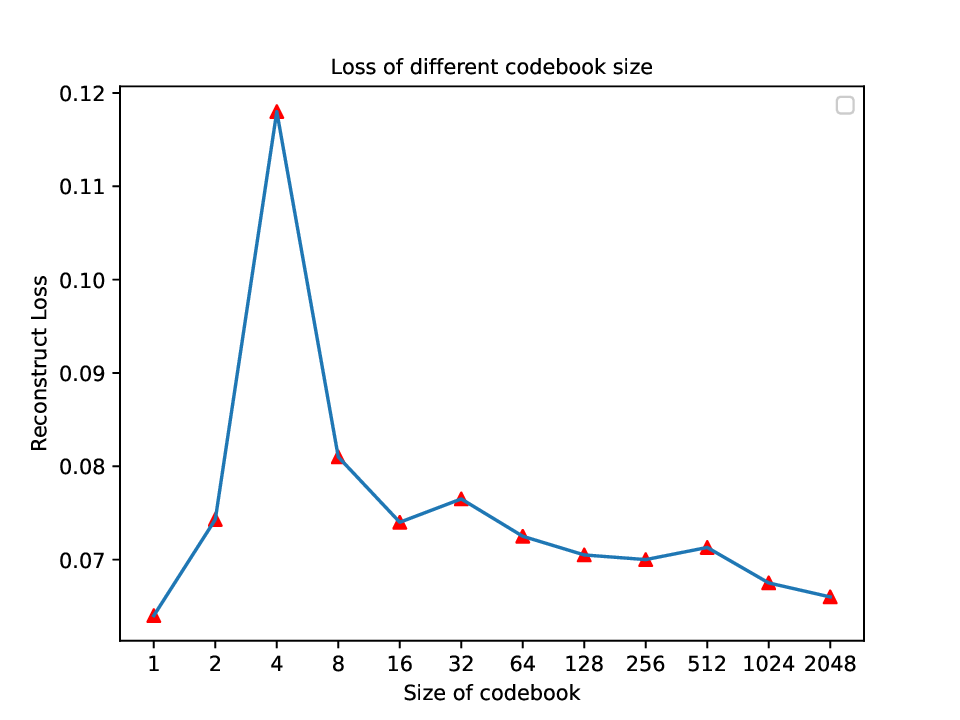}
		% \caption{图2}
		\label{图2}%文中引用该图片代号
	\end{minipage}
        % \setlength{\abovecaptionskip}{0.cm} %调整标题上方的距离
        % \setlength{\abovecaptionskip}{0.1cm} %调整标题下方的距离
        % \centering
        \caption{The communication losses under different latent space sizes. With the increase in the size of the codebook, the loss obtained by $10$ agents during the communication-validation phase shows a downward trend. Left: MNIST. Right: CIFAR10.}
        % \vspace{-0.3cm}  %调整图片与上文的垂直距离
        \label{fig9}
\end{figure}

The subsequent research will mainly focus on the usage patterns of the codebook when communicating with discrete semantics and how to improve the codebook to enhance communication efficiency. In the research, the MNIST and CIFAR10 datasets are primarily used for exploration. Firstly, We investigated the impact of the size of the latent codebook space on the efficiency of discrete communication. We conducted experiments using a single-token VQ-VAE model following Algorithm \ref{algo1}, with the number of agents $m$ set to $10$. In the experiments, we controlled the experimental variable to be the size of the first dimension of the latent space. The result is shown in Figure \ref{fig9}.
 Although there is some fluctuation in the subsequent data for the MNIST dataset, we speculate that this is due to the small dataset size and the large codebook space. Therefore, we have reason to believe that as the codebook space expands, agents can capture more patterns when learning the language, thereby further improving the efficiency of discrete communication and enhancing the performance of discrete learning.

In order to further investigate this direction in-depth, we conducted a study on the utilization of the codebook and some patterns in the VQ-VAE model. In the following results, our experiments were not conducted according to the aforementioned algorithm, but rather using a single model trained on the official datasets.
\begin{figure}[htbp]
	% \centering
        \vspace{-0.25cm}  %调整图片与上文的垂直距离
	\begin{minipage}{0.49\linewidth}
		\centering
		\includegraphics[width=0.8\linewidth]{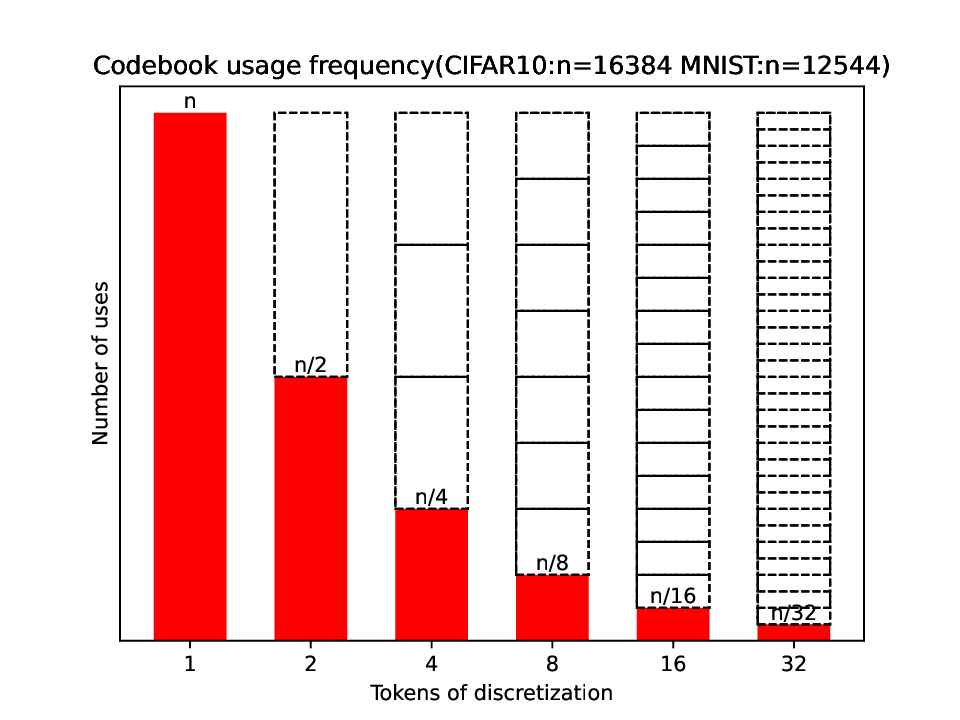}
		% \caption{图1}
		\label{图1}%文中引用该图片代号
	\end{minipage}
        \begin{minipage}{0.49\linewidth}
		\centering
		\includegraphics[width=0.8\linewidth]{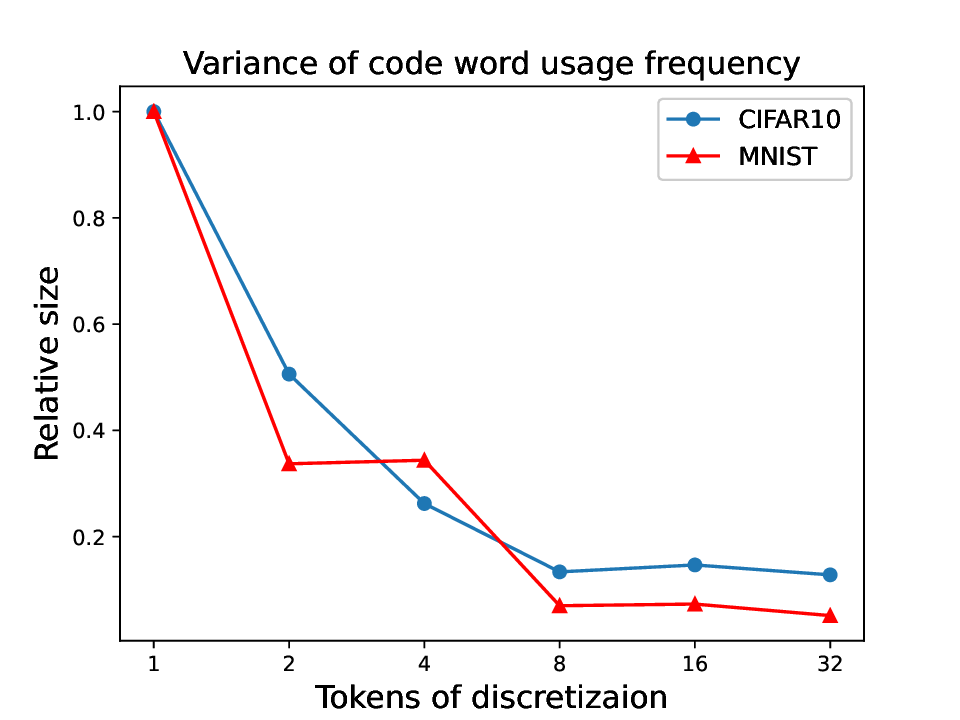}
		% \caption{图2}
		\label{图2}%文中引用该图片代号
	\end{minipage}
        \setlength{\abovecaptionskip}{0.cm} %调整标题上方的距离
        % \setlength{\abovecaptionskip}{-0.1cm} %调整标题下方的距离
        % \centering
        \caption{The pattern of codebook usage with different numbers of discrete tokens. The number of times a codebook is used strictly follows the rules based on the number of tokens, and multi-token discretization facilitates the full utilization of the codebook.}
        % \vspace{-0.4cm}  %调整图片与上文的垂直距离
        \label{fig10}
\end{figure}
Figure \ref{fig10}(Left) represents the number of times code vectors are used for each codebook update. Assuming the single-token model uses a code vector $N$ times for each codebook update. For an $m$-token model, each codebook update occurs $\frac{N}{m}$times. In each iteration, the codebook is updated $m$ times, so after implementing multi-token discretization, the codebook updates strictly follow the rule based on the number of tokens, with the total usage of code vectors in each iteration remaining $N$, and any $m$-token model updating the codebook $m$ times within that iteration, each token using $\frac{N}{m}$ code vectors. Figure \ref{fig10}(Right) represents the variance between the frequencies of use of different code words for different numbers of discrete tokens. It can be observed that as the number of discrete tokens increases, the codebook is utilized more evenly.
\begin{figure}[htbp]
	% \centering
        % \vspace{-0.45cm}  %调整图片与上文的垂直距离
	\begin{minipage}{0.48\linewidth}
		\centering
		\includegraphics[width=0.8\linewidth]{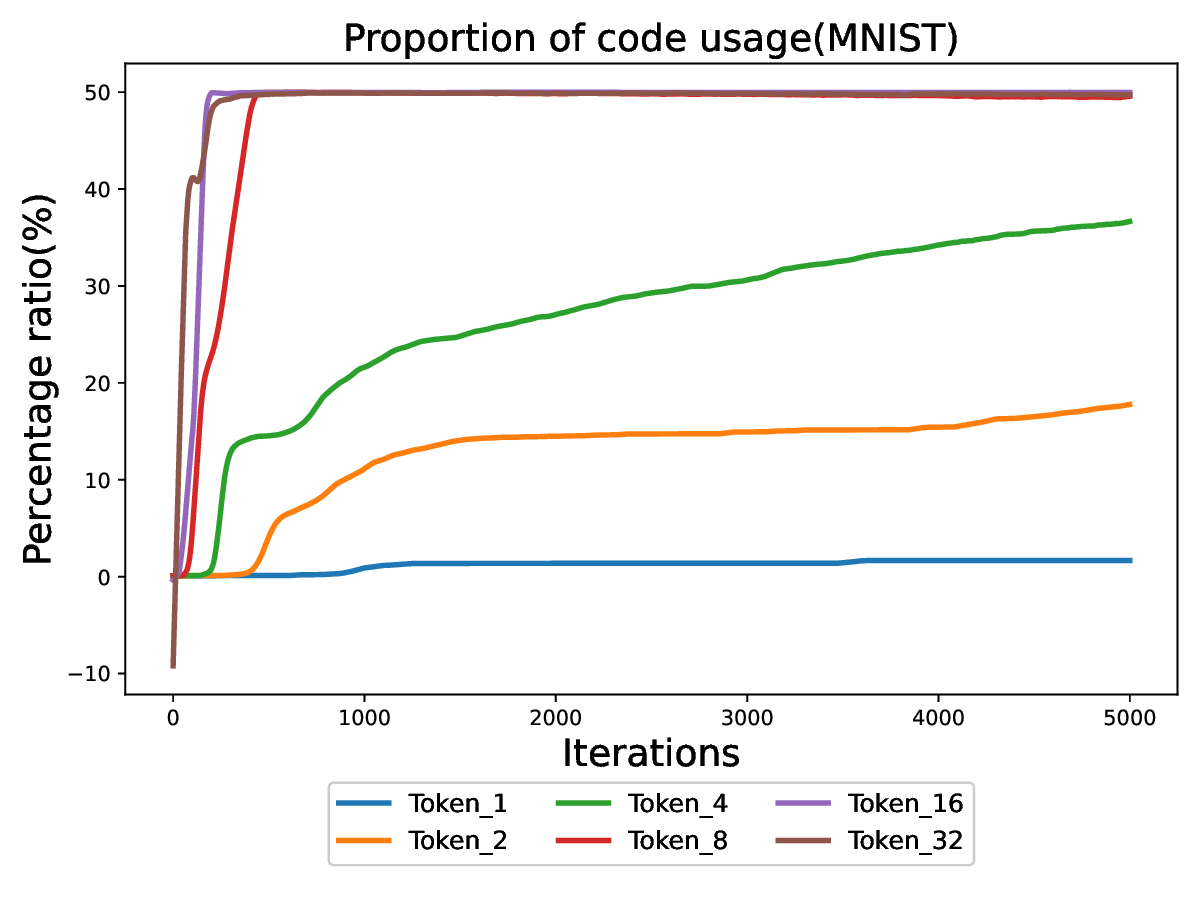}
		% \caption{图1}
		\label{图1}%文中引用该图片代号
	\end{minipage}
        \begin{minipage}{0.48\linewidth}
		\centering
		\includegraphics[width=0.8\linewidth]{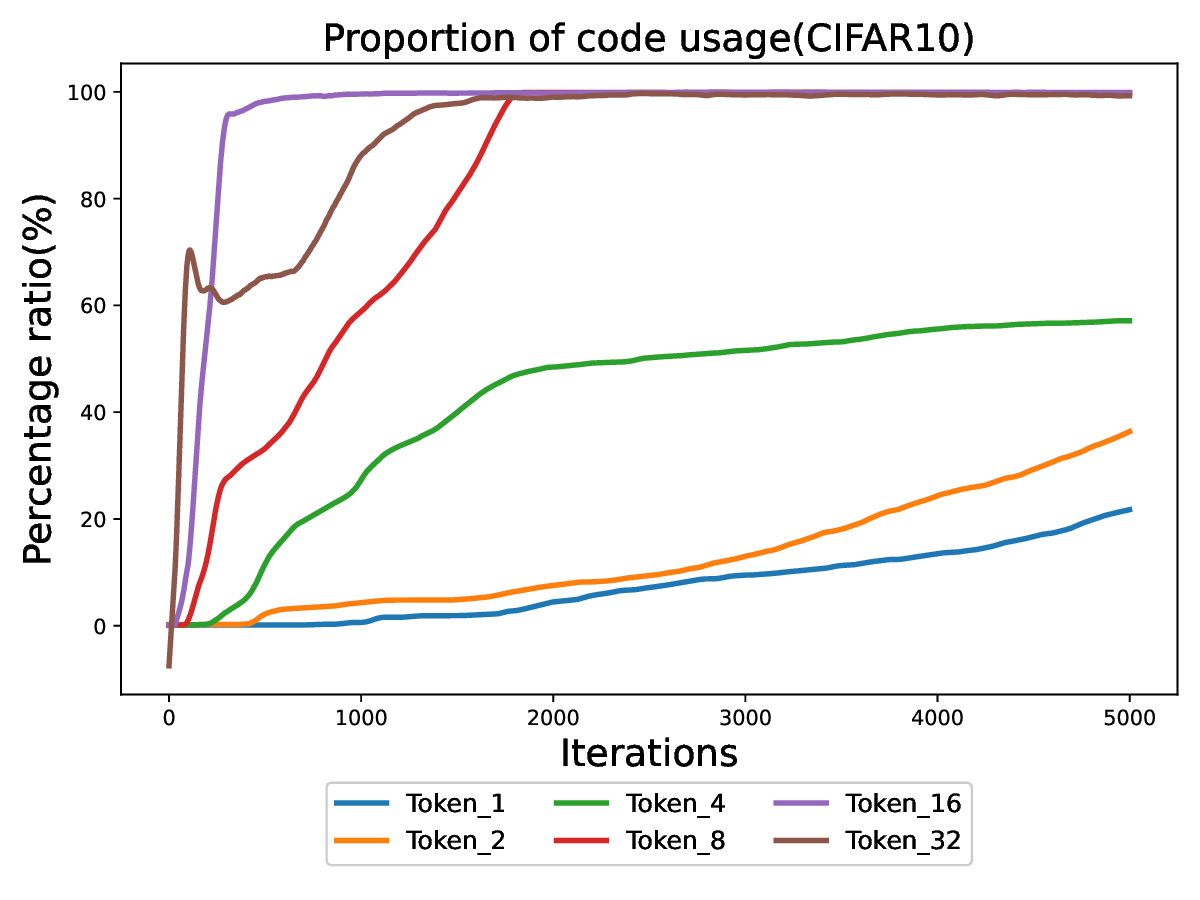}
		% \caption{图2}
		\label{图2}%文中引用该图片代号
	\end{minipage}
        \setlength{\abovecaptionskip}{0.cm} %调整标题上方的距离
        % \setlength{\abovecaptionskip}{-0.1cm} %调整标题下方的距离
        % \centering
        \caption{Utilization rate of the codebook during the training iterations. Multi-token discretization facilitates the full utilization of the codebook.}
        % \vspace{-0.4cm}  %调整图片与上文的垂直距离
        \label{fig11}
\end{figure}
Figure \ref{fig11} shows the proportion of the number of codewords used in each iteration to the total number, under different numbers of discrete tokens. It can be observed that when the number of discrete tokens is greater than or equal to $8$, the codebook is effectively utilized throughout the iterations. 
\begin{figure}[htbp]
	% \centering
        \vspace{-0.25cm}  %调整图片与上文的垂直距离
	\begin{minipage}{0.48\linewidth}
		\centering
		\includegraphics[width=0.8\linewidth]{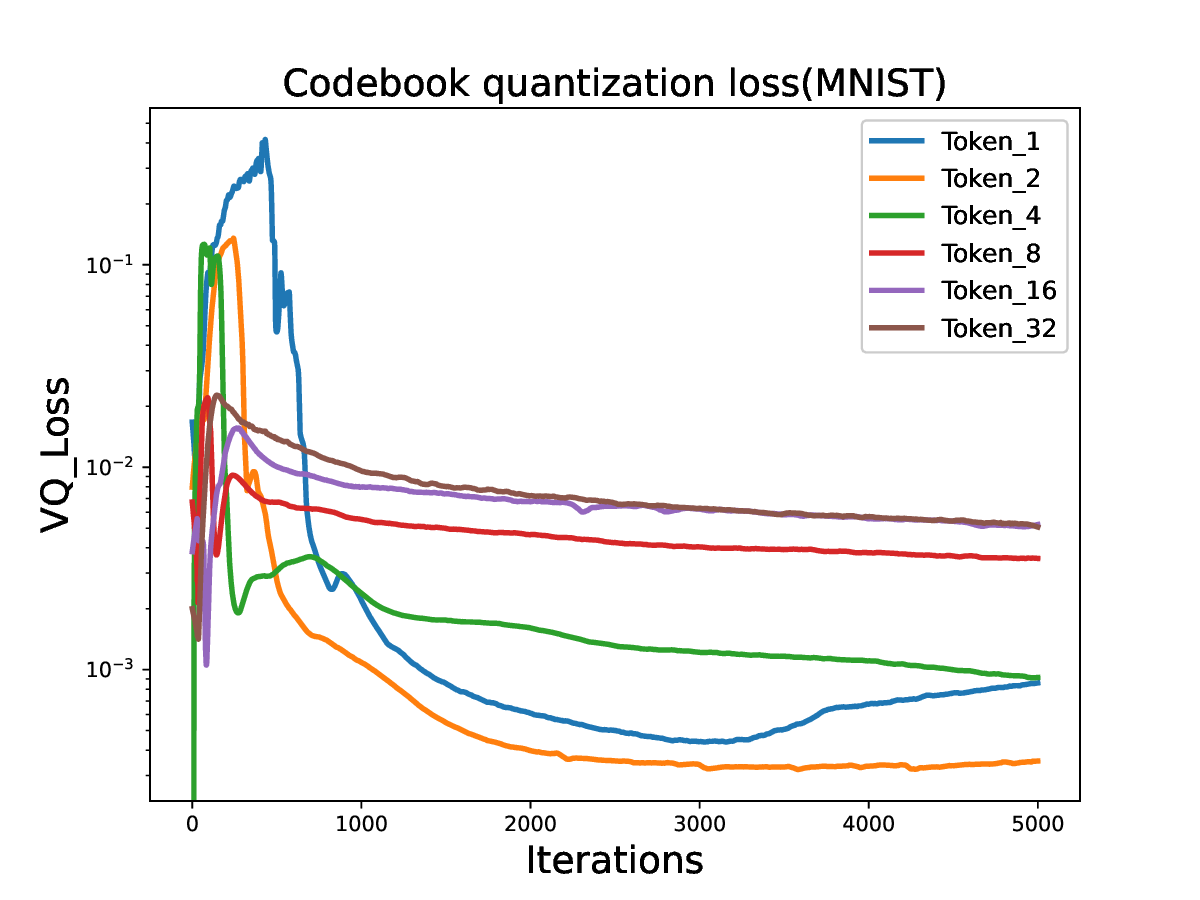}
		% \caption{图1}
		\label{图1}%文中引用该图片代号
	\end{minipage}
        \begin{minipage}{0.48\linewidth}
		\centering
		\includegraphics[width=0.8\linewidth]{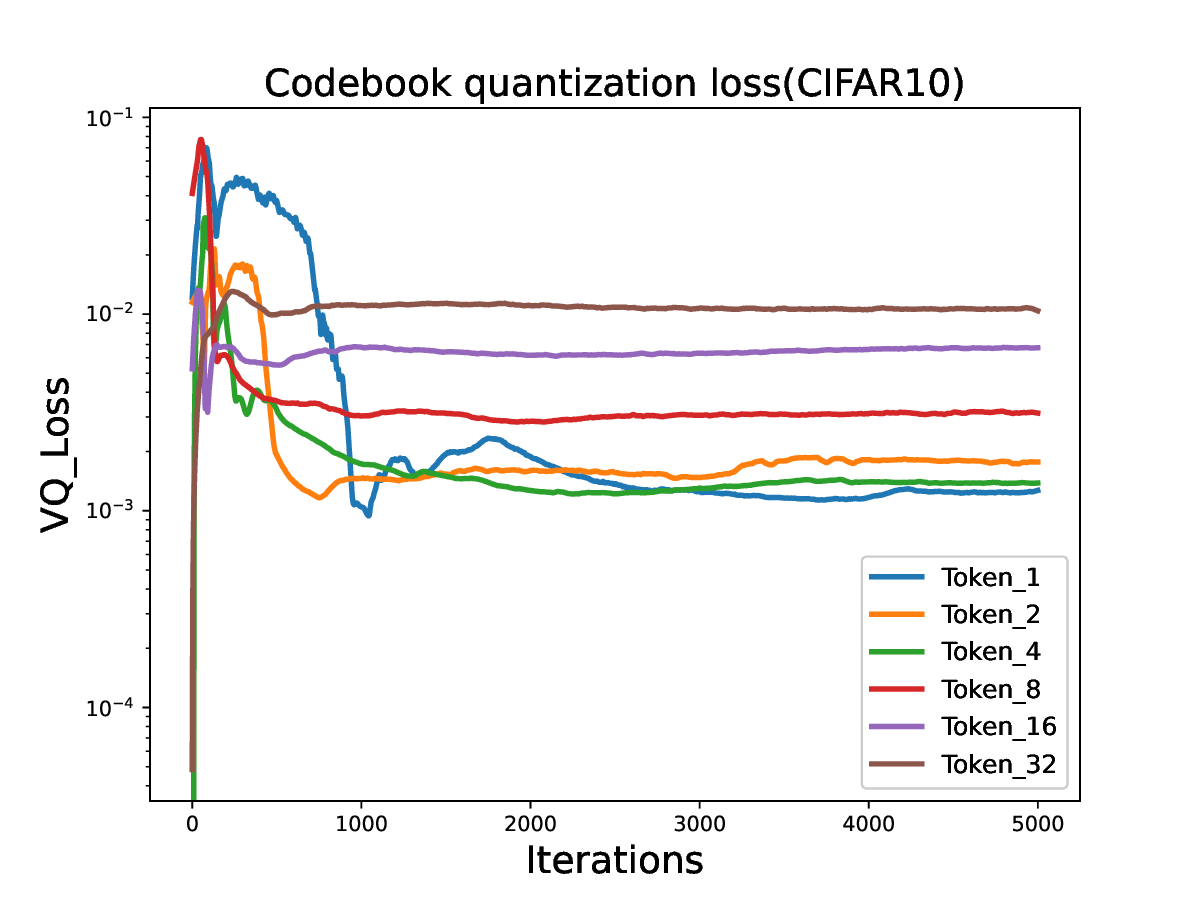}
		% \caption{图2}
		\label{图2}%文中引用该图片代号
	\end{minipage}
        \setlength{\abovecaptionskip}{0.cm} %调整标题上方的距离
        % \setlength{\abovecaptionskip}{-0.1cm} %调整标题下方的距离
        % \centering
        \caption{Codebook quantization loss under different numbers of discrete tokens during training. Although multi-token discretization may cause an increase in quantization loss, the codebook is evidently learned more evenly.}
        % \vspace{-0.4cm}  %调整图片与上文的垂直距离
        \label{fig12}
\end{figure}
Figure \ref{fig12} shows the transformation of codebook quantization loss for different numbers of discrete tokens. As the number of training iterations increases, the codebook's quantization loss gradually stabilizes, and a higher number of discrete tokens results in a higher stable loss value.
\begin{figure}[htbp]
	% \centering
        % \vspace{-0.55cm}  %调整图片与上文的垂直距离
        \centering
        \includegraphics[width=0.32\linewidth]{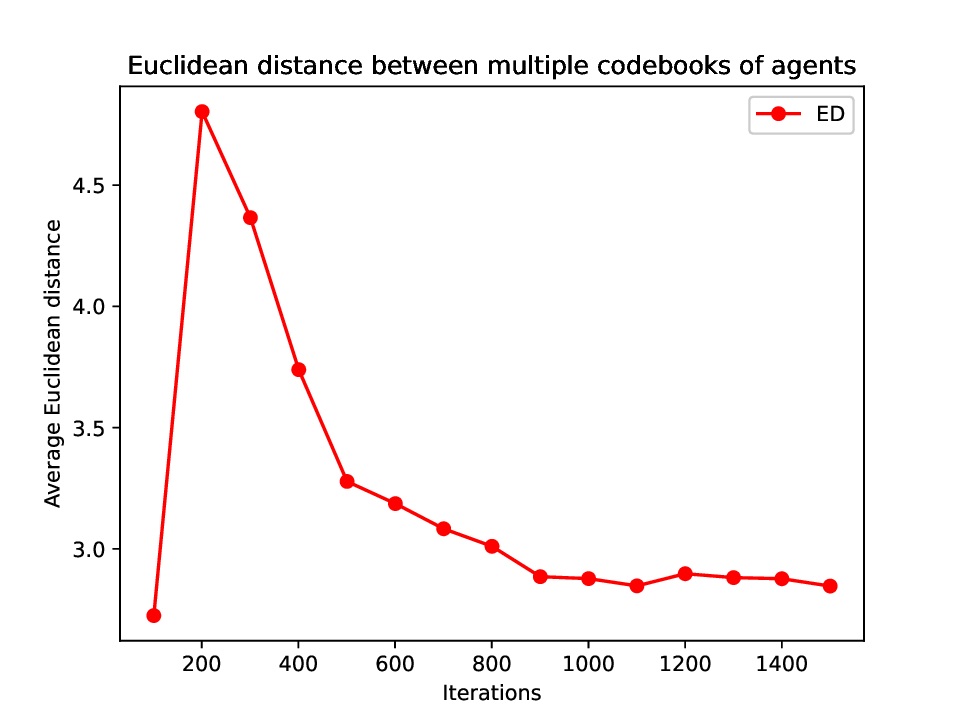}
        \hfill
        \includegraphics[width=0.32\linewidth]{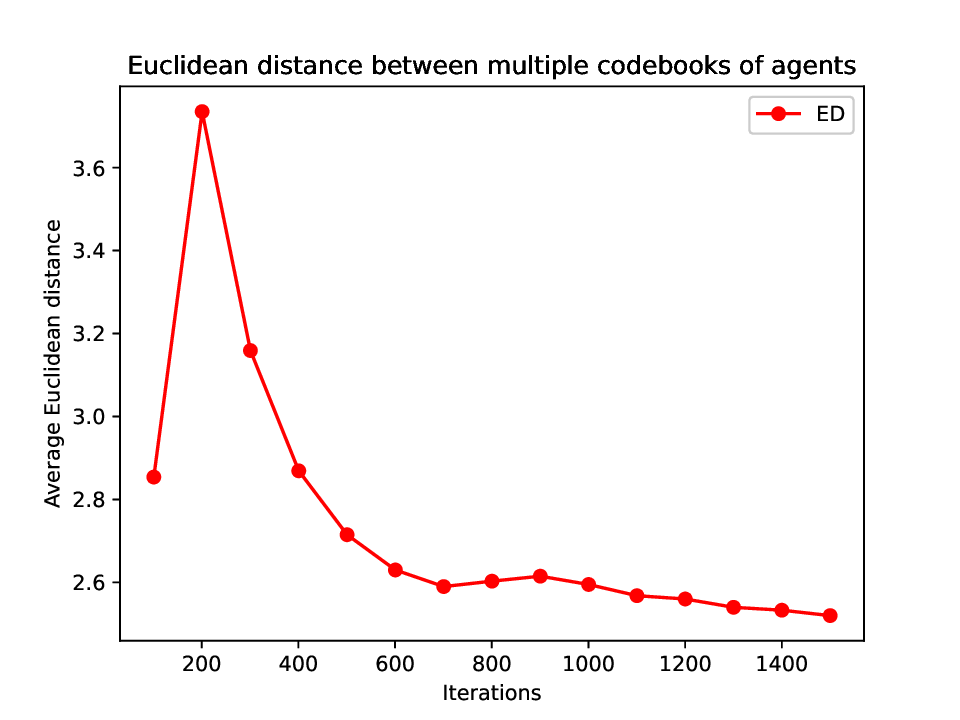}
        \hfill
        \raisebox{2mm}{\includegraphics[width=0.32\linewidth]{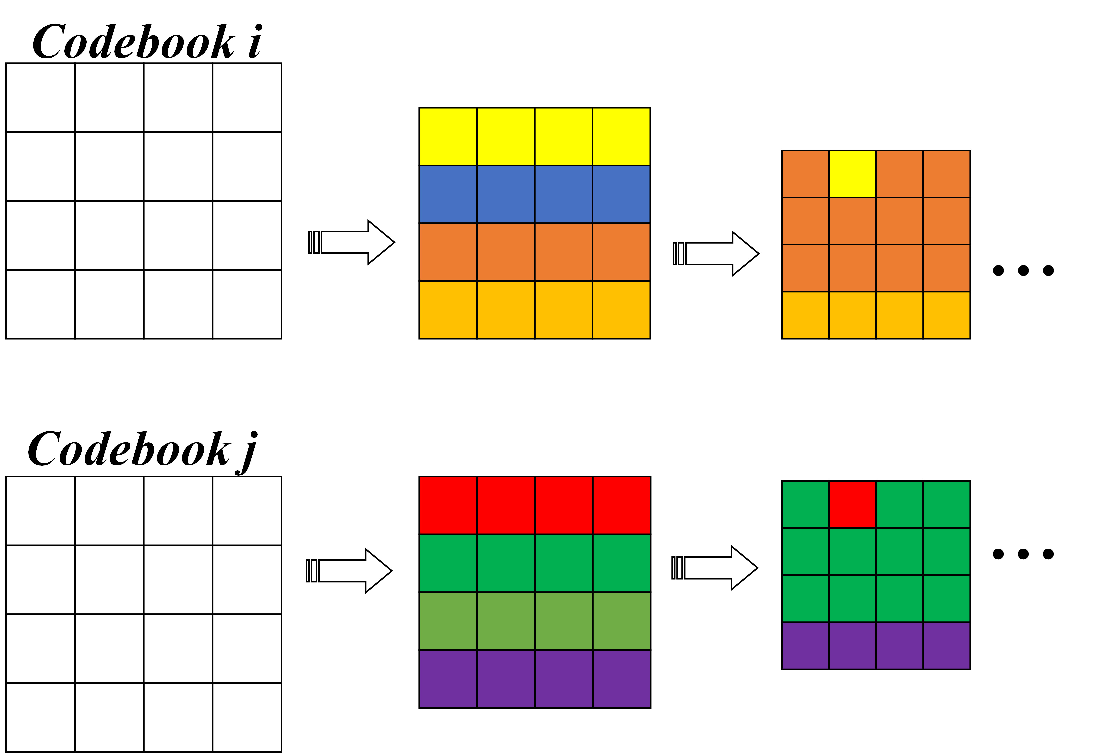}}
        \setlength{\abovecaptionskip}{0.cm} %调整标题上方的距离
        \setlength{\abovecaptionskip}{-0.1cm} %调整标题下方的距离
        % \centering
        \caption{The extent of difference in understanding of the same language among $10$ agents. As learning progresses, their similarity in language comprehension increases. Left: Experiments on MNIST. Mid: Experiments on CIFAR10. Right: The diagram to show that different codebooks become more and more similar.}
        % \vspace{-0.4cm}  %调整图片与上文的垂直距离
        \label{fig13}
\end{figure}
For $m$ agents with an overlap rate of $0.1$, we explore the codebook similarity among them during the training process, using the Euclidean distance as a measure. Assuming a codebook size of $\mathbb{R}^{L\times M }$, the calculation of the Euclidean distance is shown in Equation \ref{eq5}.
\begin{equation}
\label{eq5}
\mathbb{ED}_{Average}  = \frac{1}{{m(m-1)/2}}\sum_{i=1}^{m}\sum_{j=i+1}^{m}\sqrt{\sum_{u=1}^{L}\sum_{v=1}^{M}(C_{i} (u,v) - C_{j}(u,v))^2}
\end{equation}
$C_{i},C_{j}$ represents the codebook of different agents.
Figure \ref{fig13} illustrates the Euclidean distances between pairwise codebooks of ten agents during the learning process, with an overlap of $0.1$. As the iterative learning progresses, the Euclidean distances between the latent codebooks of different agents decrease, indicating an increase in their similarity. 

In section \ref{subsec3} of the research content, it is stated that increasing the size of the latent codebook space is beneficial for agents to improve their discrete communication efficiency. The patterns of codebook usage will aid in our future research endeavors. Especially the multi-token discretization mechanism, which has improved the issue of uneven codebook usage and mitigated the discretization bottleneck.

\section{Conclusion, Limitation and Future Study}\label{sec5}
Our experiments have shown that communication between agents using single-token discrete semantics can achieve comparable efficacy to that of continuous semantics. However, multi-token discretization before communication significantly enhances the quality of information exchange compared to continuous language. Additionally, multi-token discretization outperforms single-token approaches in terms of system generalization. This suggests that multi-token communications are more effective, especially in contexts where agents encounter diverse languages. Furthermore, we have explored the use of a VAE model versus an AE model for communication, and found that agents' communication facilitated by the VAE model outperforms VQ-VAE model-based communication. We aim to investigate the underlying reasons for this discrepancy in future research. In summary, the multi-token discretization approach we propose outperforms the original single-token discretization method, and compared to continuous communication based on the AE model, using multi-token discretization offers a greater advantage for communication among isolated agents.
\section*{Acknowledgments}
We would like to express our deep gratitude to the authors of the original model VQ-VAE. Their model provides valuable foundation and inspiration for our research. In addition, we would like to thank other relevant researchers for providing code. URL: https://github.com/zalandoresearch/pytorch-vq-vae?tab=MIT-1-ov-file.

\appendix
\section{Appendix}\label{appendix}
The models used in this paper were implemented on the PyTorch $1.12.1$ framework, using PyCharm Community Edition $2023.1$ on the Windows platform. The model training was conducted on a single GeForce RTX $3060$ GPU with 8GB of GPU memory, using the CUDA $12.1$ experimental environment. The operating system used was Windows. Here we provide additional details about the experimental setup and additional results. 

\begin{figure}[htbp]
    \vspace{-0.55cm}  %调整图片与上文的垂直距离
    \centering
    \includegraphics[width=0.45\linewidth]{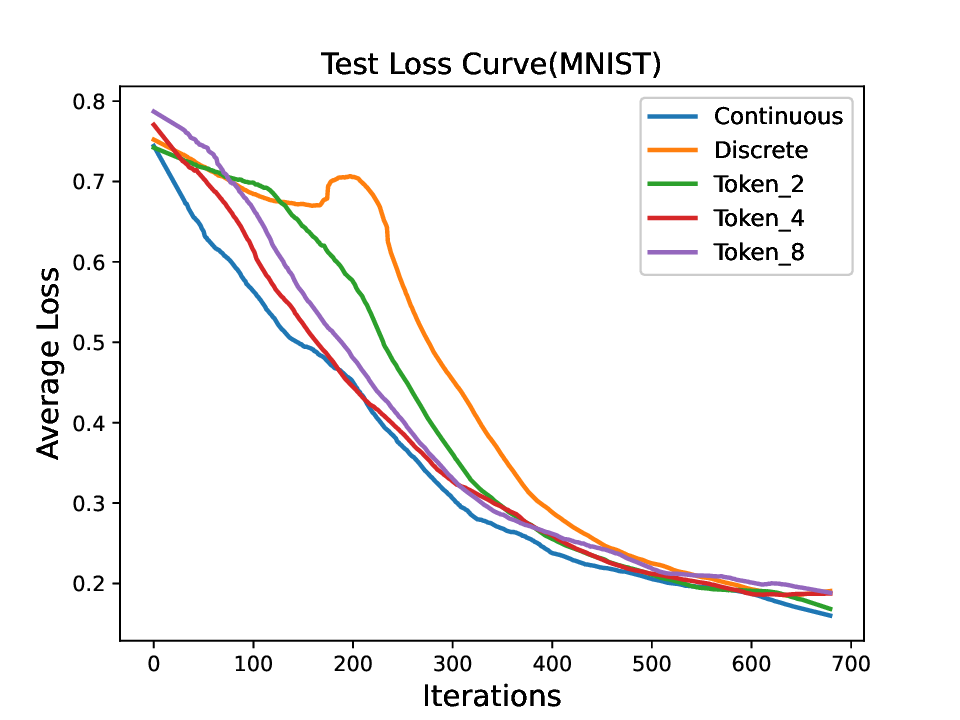}
    \hfill
    \includegraphics[width=0.45\linewidth]{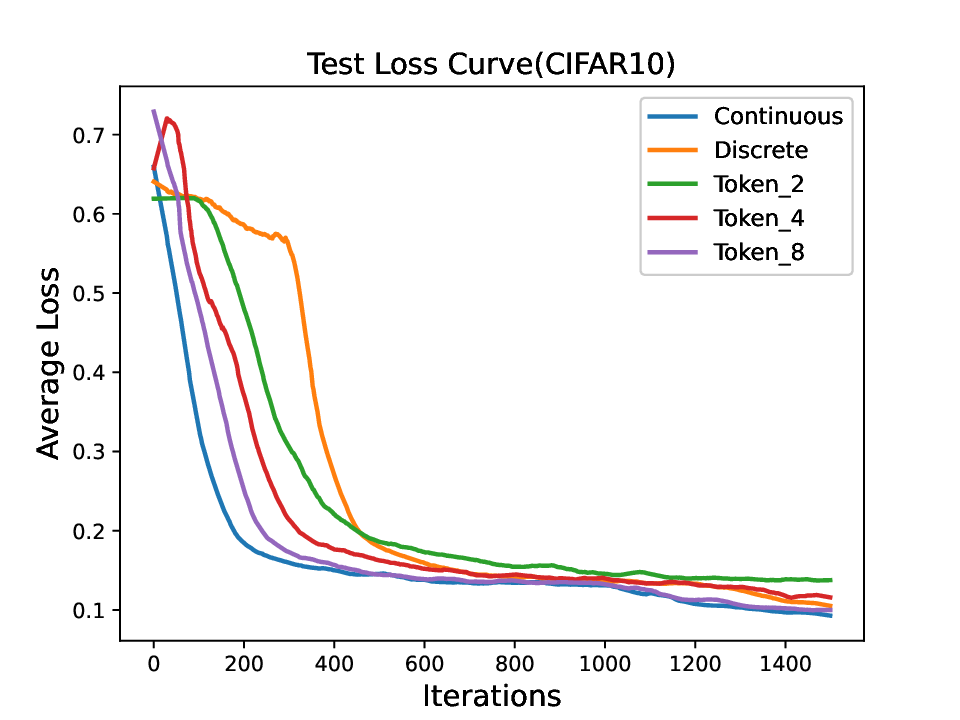}
    \\ % 换行
    % 第二行
    \includegraphics[width=0.45\linewidth]{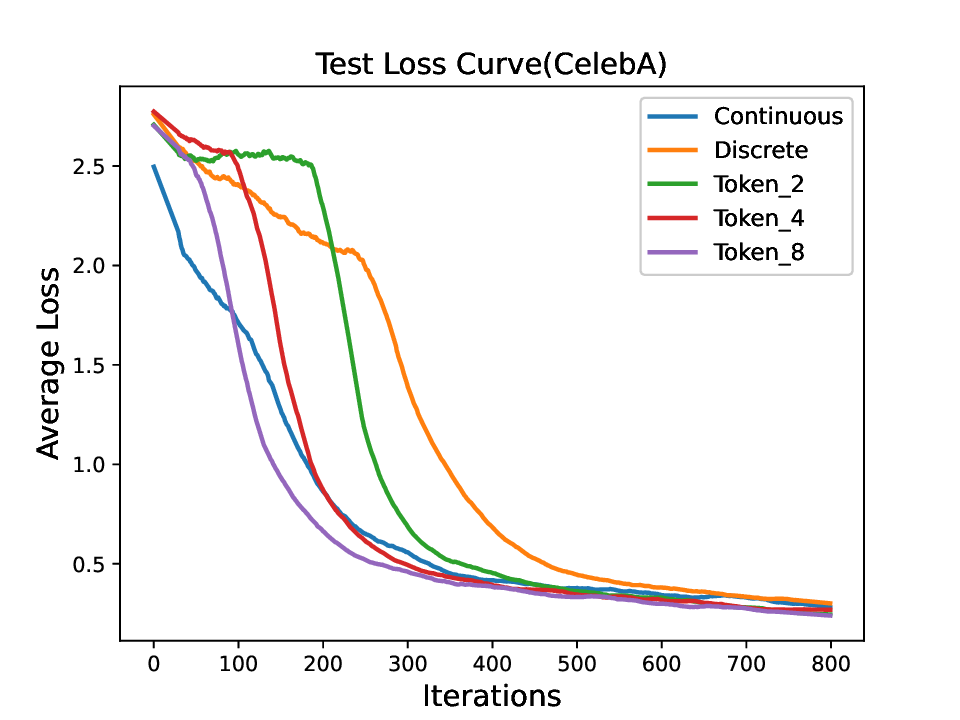}
    \hfill
    \includegraphics[width=0.45\linewidth]{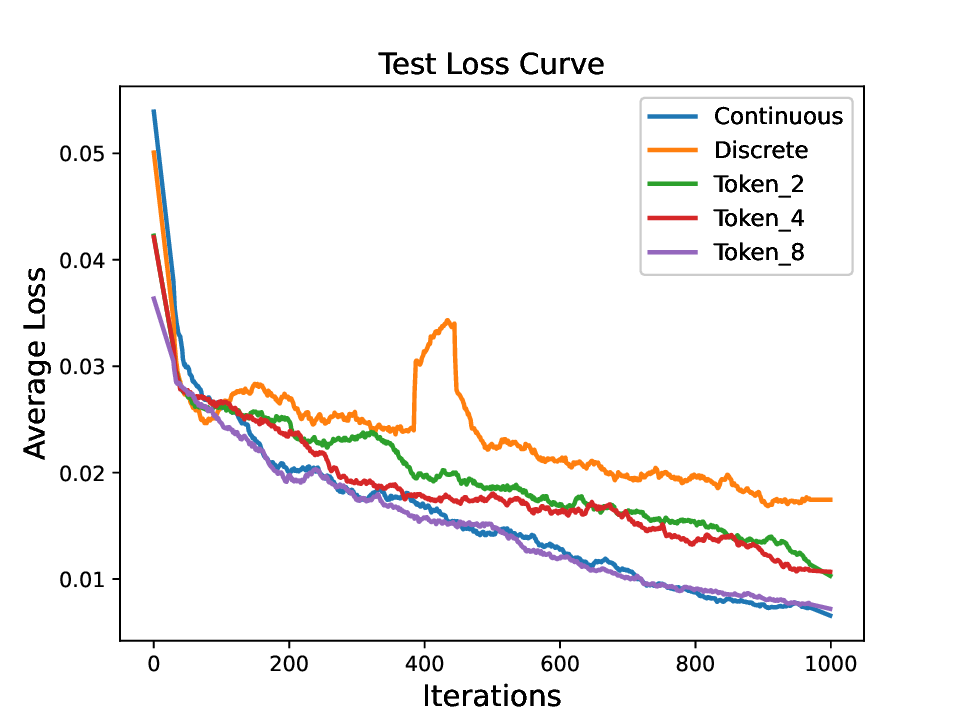}
    \setlength{\abovecaptionskip}{-0.1cm} %调整标题下方的距离
    \caption{Internal test loss during cross-training of multiple agents. Compared to the single token, multi-token discretization has improved the learning speed of agents for communication languages.}
    % \vspace{-0.4cm}  %调整图片与上文的垂直距离
    \label{fig14}
\end{figure}

% \vspace{0.3cm}
\begin{algorithm}

\caption{Cross-training Process}\label{algo2}

\begin{algorithmic}[1]
% \begin{algorithmic}[1]
\State {Train $m$ agents simultaneously;}

% \Ensure Train $m$ pairs of agents simultaneously.

\State {Initialize encoders $E = \{e_0, ..., e_{m}\}$}

\State {Initialize quantization layers $H = \{h_0\left ( \cdot  \right ), ..., h_{m}\left ( \cdot  \right )\}$}

\State {Initialize decoders $D = \{d_0, ..., d_{m}\}$}

\For {each iteration $i$}
  
    \State {Sample input data $x$}
    
    \State {Randomly sample encoder $e_i$, $d_k$, $c$($i\neq k$, $c$ is $c_i$ or $c_k$ )}

    \State {$z_{q} \Leftarrow e_{i}(x)$}

    \State {$z_{i},L_{quantify} \Leftarrow h_{i}(z_{q} )$}

    \State {$x' \Leftarrow d_{i}(z_{i} )$}

    \State {$L_{i} \Leftarrow MSE(x',x) + L_{quantify}  $}

    \State {optimize $e_{i}$, $h_{i}$ and $d_{i}$ with respect to $L_{i}$}

    \For {each pair of agents $j$}
   
    \State {Sample validate data $v$}

    \State {$loss_{j} \Leftarrow d_{j} (h_{j}(e_{j}(v) ) ) $}
  
    \State {output $loss_{j}$} 
    
  \EndFor
  
\EndFor

\end{algorithmic}
\end{algorithm}
% \vspace{-0.3cm}
The CelebA dataset we used in all experiments was $8000$ images extracted from official sources, and in the experiment of Algorithm \ref{algo1}, we divided them into $8$ categories based on their attributes for use by $8$ agents. In Figure \ref{fig6}, we have demonstrated that the multi-token discretization mechanism is more effective in terms of communication between agents compared to the single-token discretization mechanism. Prior to this, we had already conducted some experimental work to prove the feasibility of the multi-token discretization mechanism. Algorithm \ref{algo2} is the process through which we validate the effectiveness of multi-token discrete communication. For the dataset used in Algorithm \ref{algo2}, MNIST and CIFAR10 are used directly without any changes, while the CelebA and retinal datasets are divided into two categories: training and validation sets. That is, the agents in the algorithm are exposed to the same training or validation sets. We varied the intermediate processing architecture between the speaker and listener and recorded the test loss of $m$ agents throughout the entire training process, as shown in Figure \ref{fig14}. 

The curves in the figure indicate that when using cross-training, multi-token discretization indeed outperforms the single-token approach. Furthermore, the results demonstrate a pattern where increasing the number of tokens leads to better performance and faster learning. That's why we initially wanted to use multi-token discretization mechanism for agents' communication. AE model still exhibits the fastest learning speed. That is, when agent learn a language, those that adopt a continuous semantic approach learn the fastest. However, as shown in Figure \ref{fig5}, when these agents interact with new agents, the outcomes are not as good as those of agents that learned through a discrete semantic approach.
\begin{figure}[htbp]
	\centering
        % \vspace{-0.4cm} 
	\includegraphics[width=0.8\linewidth]{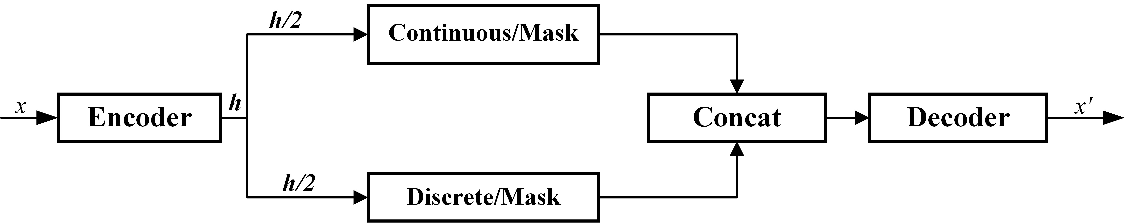}
        \setlength{\abovecaptionskip}{0.1cm} %调整标题上方的距离
        % \setlength{\abovecaptionskip}{cm} %调整标题下方的距离
        % \centering
        \caption{Different methods for communication between agents. One method is to perform continuous communication on half of the information and discrete communication on the other half. Another method is to mask half of the information and use the remaining half for continuous or discrete communication. Here, two models are used to simulate continuous communication, namely autoencoder (AE) and variational autoencoder (VAE).}
        % \vspace{-0.45cm}  %调整图片与上文的垂直距离
        \label{fig15}
\end{figure}
In section \ref{sec4}, we mentioned the configuration of experimental parameters. For the four datasets, we adjusted the batch size or the size of the latent code space accordingly. However, all parameters for experiments within the same dataset must remain consistent. During our experiments, we attempted to use the VAE model instead of the AE model as a baseline, and simulated the learning and communication process between a pair of agents using continuous semantics. Similarly, we used Equation \ref{eq4} to allocate individual datasets to each agent, and the loss during training of the VAE model is represented by Equation \ref{eq6}.
\begin{multline}
\label{eq6}
\mathcal{L} _{VAE}= \mathcal{L} _{recon}+  \beta \mathcal{L} _{KL} \\
= \left \| x-x' \right \| _{2}+  \frac{\beta }{2}\sum_{j=0}^{J}\left ( 1+  log\left (  \sigma _{j}^{2} \right )- \mu  _{j}^{2}-\sigma _{j}^{2} \right )
\end{multline}

The first term represents the reconstruction loss, the second term represents the KL divergence loss, which characterizes the difference between the actual distribution of variables in the latent space and the prior distribution (usually assumed to be a standard normal distribution). Here, $\mu$ and $\sigma$ respectively denote the mean and standard deviation of this distribution, while $\beta$ represents a hyperparameter. 

Regarding all the experiments on the autoencoder (AE) model, we replaced it with a variational autoencoder (VAE) model and repeated the experiments. When repeating the core experiments of Algorithm \ref{algo1}, we found that the agents learning with discrete variables did not achieve very good results when communicating with each other. That is, the loss from communication was relatively high. In contrast, the agents using continuous semantics for communication showed higher efficiency in their exchanges, with lower communication loss. 

To further explore and compare the performance of continuous semantic communication based on the VAE model and discrete semantic communication, we have devised a series of experimental setups (see Figure \ref{fig15}). According to these setups, we conducted experiments, where the first major category of experiments involved the speaker's output being processed discretely for half of the information and the other half either being processed continuously. The second major category involved one half of the information being processed either discretely or continuously, while the other half was masked as zero.
We have conducted extensive experiments on the structure of Figure \ref{fig15}, but still have not yielded good results. The experimental results indicate that regardless of whether agents are trained and learned through the aforementioned combined model, or learned with singel model by masking half, under the condition of a lower overlap ratio, the relationship between the three types of cross-validation losses is: $AE> VQVAE> VAE$

However, this pattern is not absolute. Our main evaluation metric is the reconstruction loss between unfamiliar agents, as we mentioned in Section \ref{sec3}, where unfamiliarity indicates that their training datasets are not completely identical. When the proportion of overlap in the datasets is low, the performance of discrete communication methods is indeed inferior to continuous communication methods based on VAE.
\begin{figure}[htbp]
    \vspace{-0.54cm}  %调整图片与上文的垂直距离
    \centering
    % 第一行
    \includegraphics[width=0.45\linewidth]{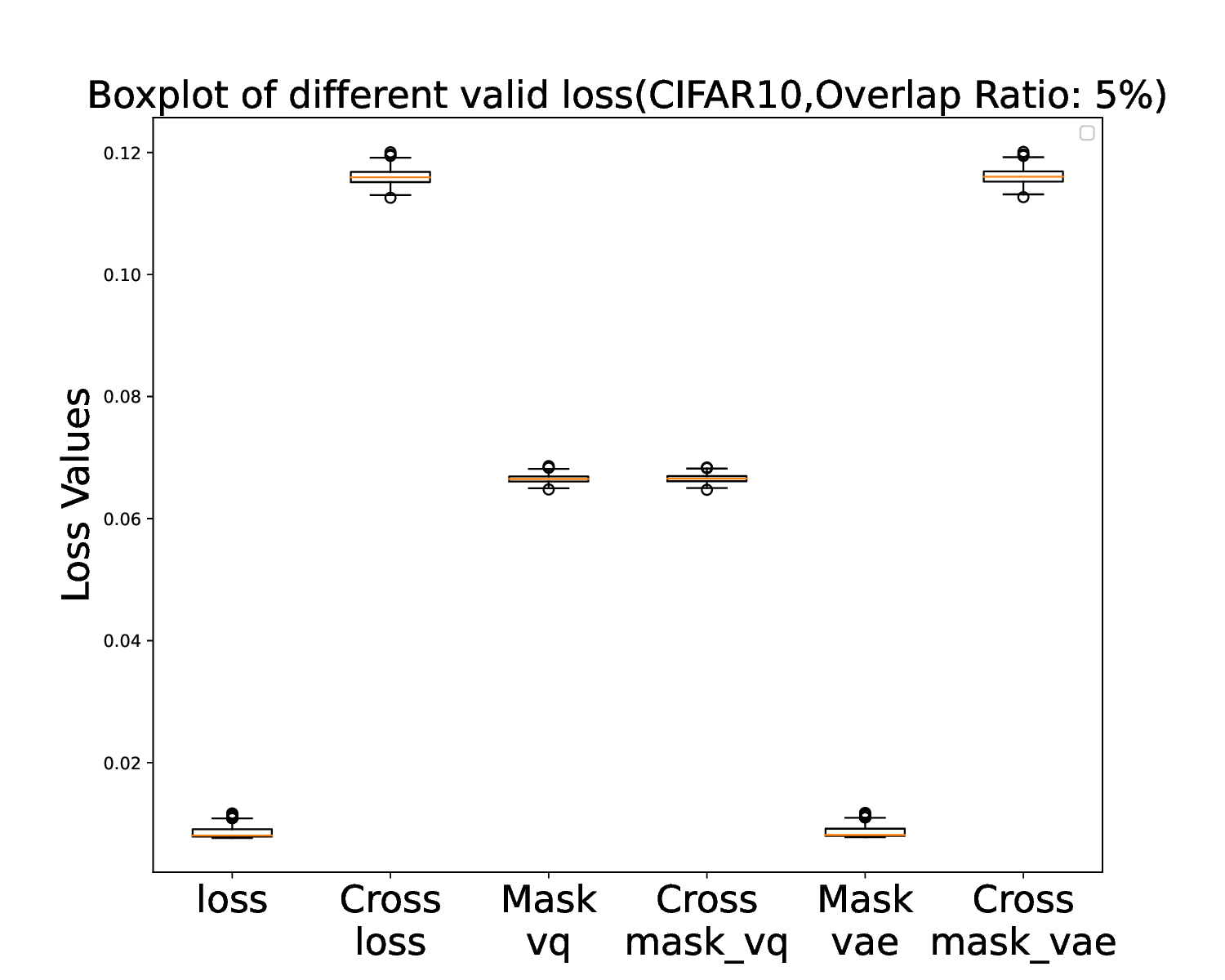}
    \hfill
    \includegraphics[width=0.45\linewidth]{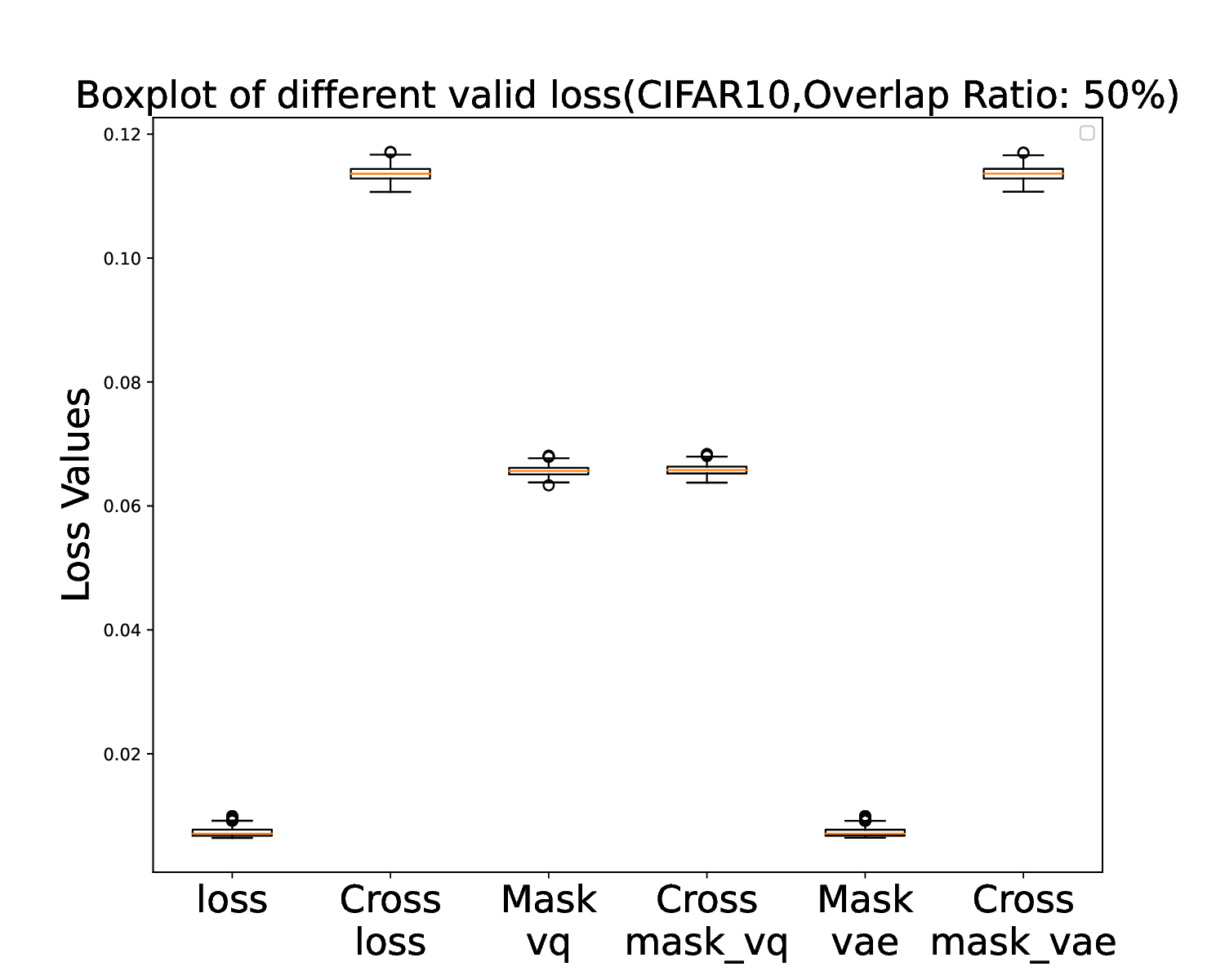}
    \\ % 换行
    % 第二行
    \includegraphics[width=0.45\linewidth]{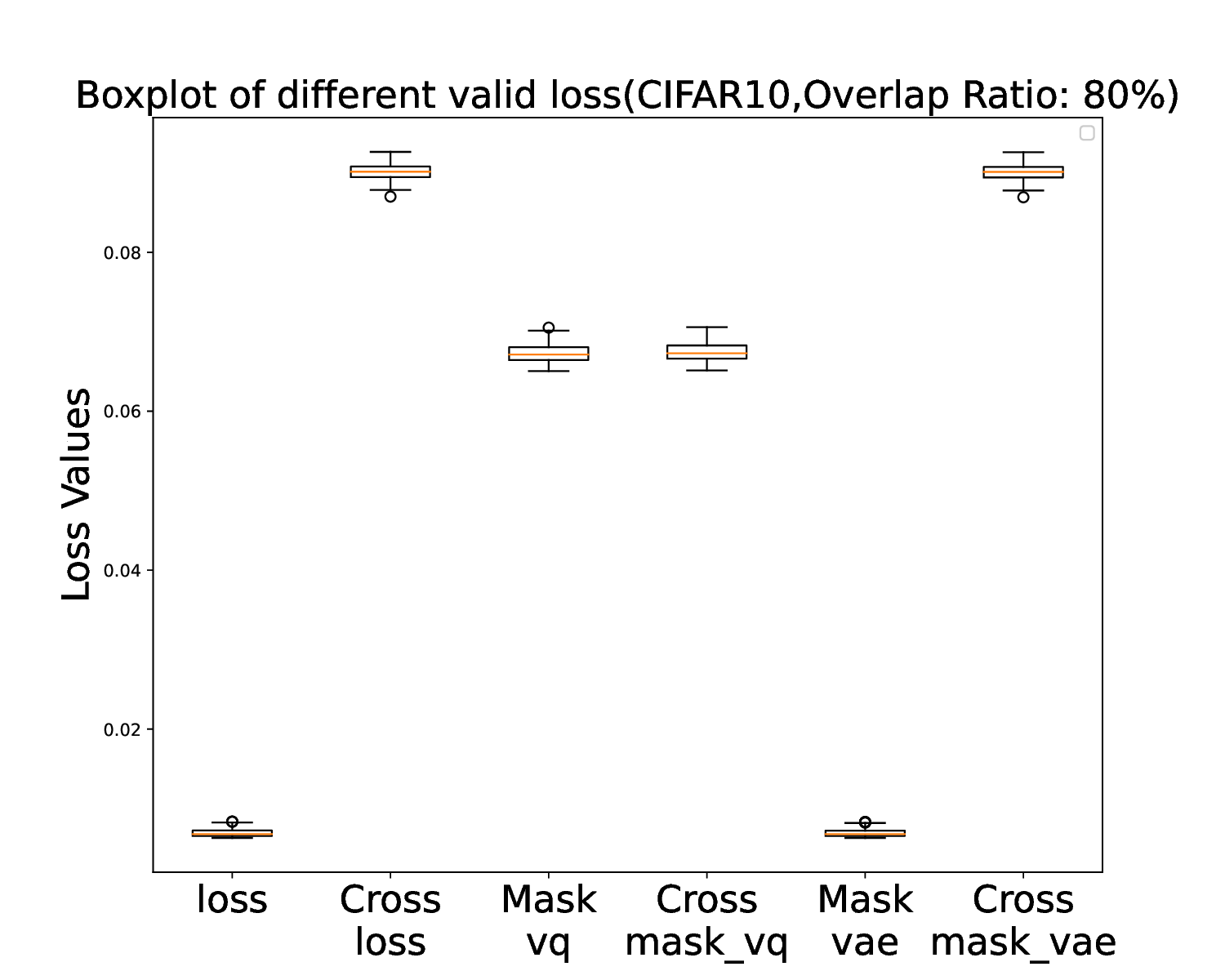}
    \hfill
    \includegraphics[width=0.45\linewidth]{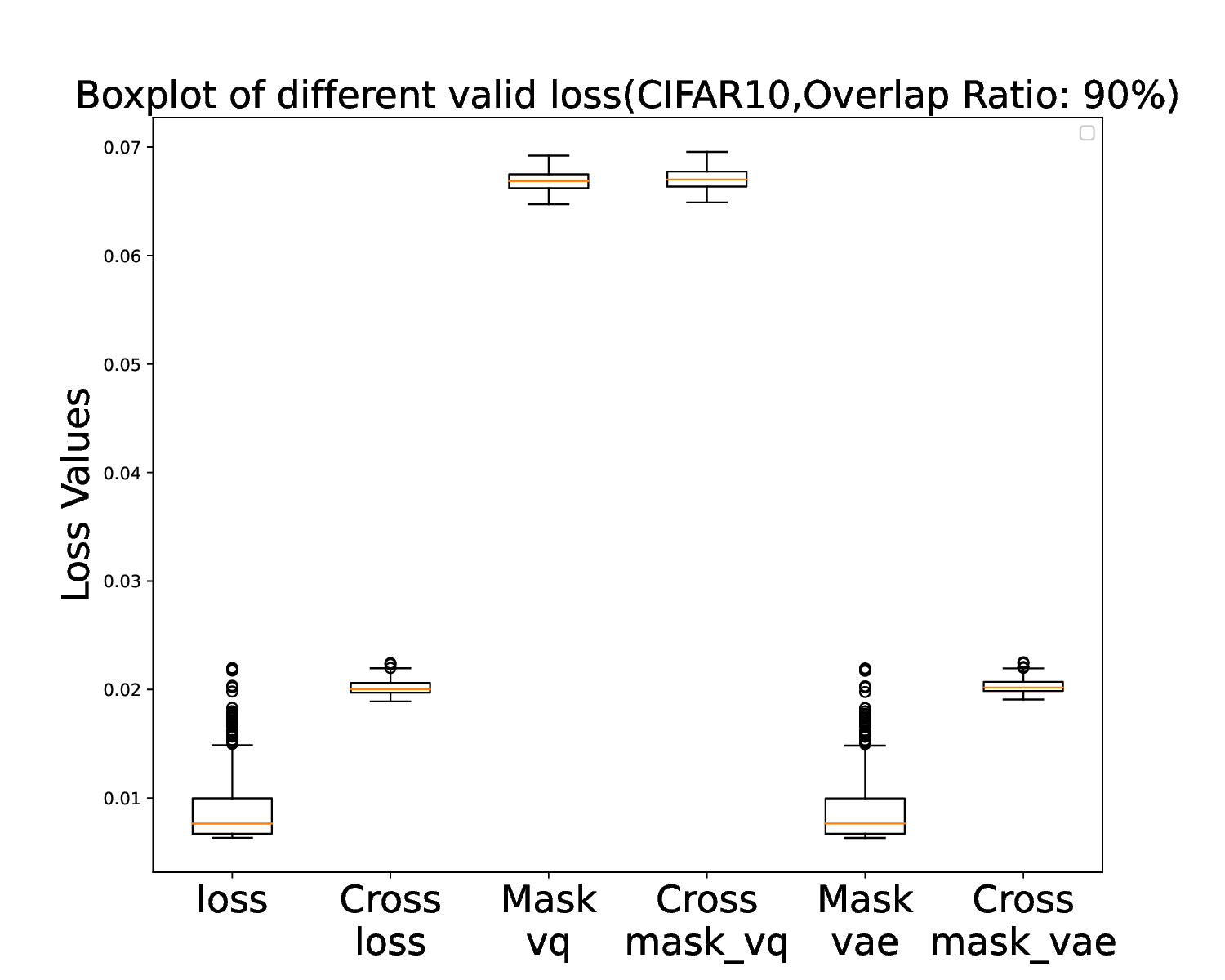}
    \\ % 换行
    % 第三行
    \includegraphics[width=0.45\linewidth]{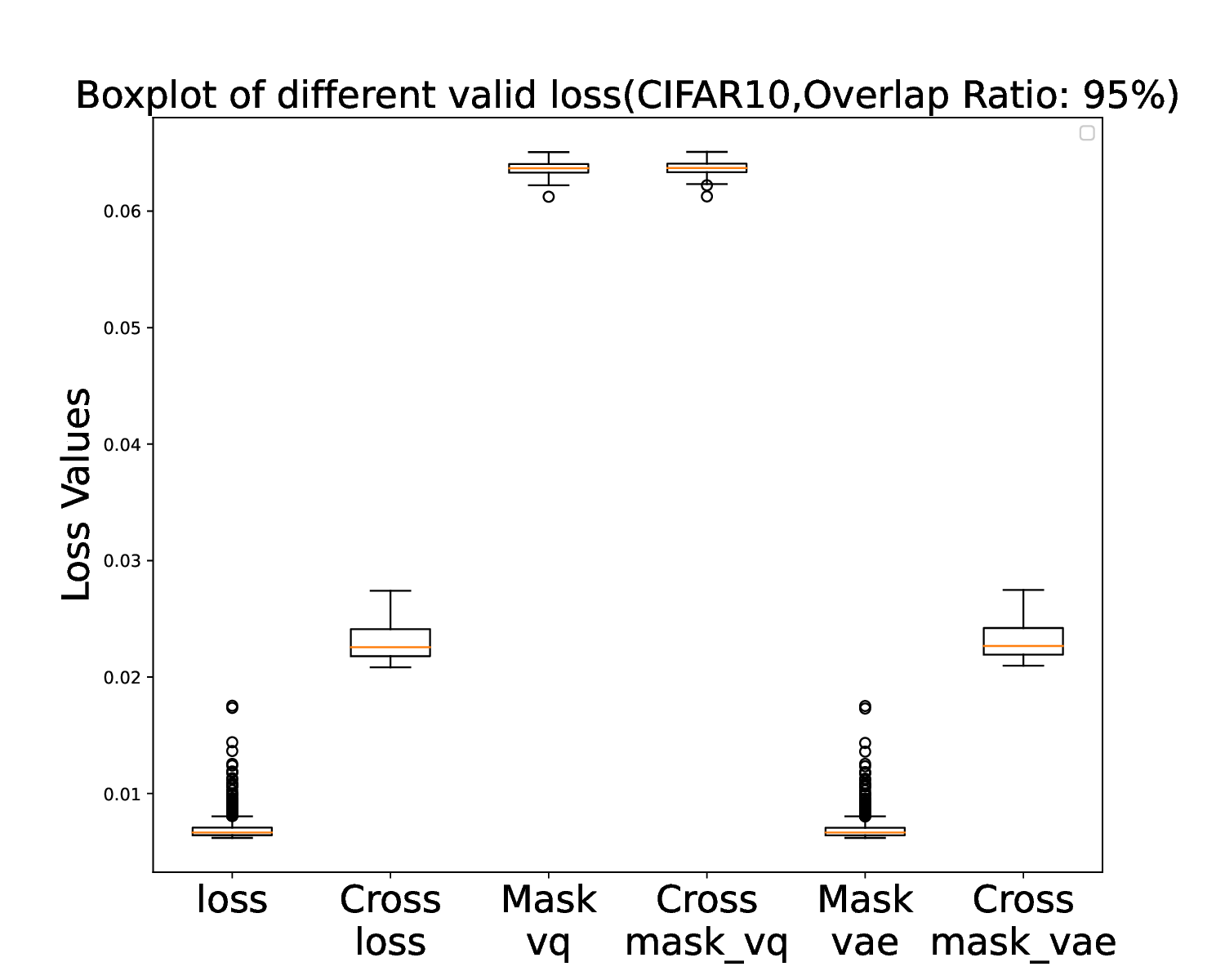}
    \hfill
    \includegraphics[width=0.45\linewidth]{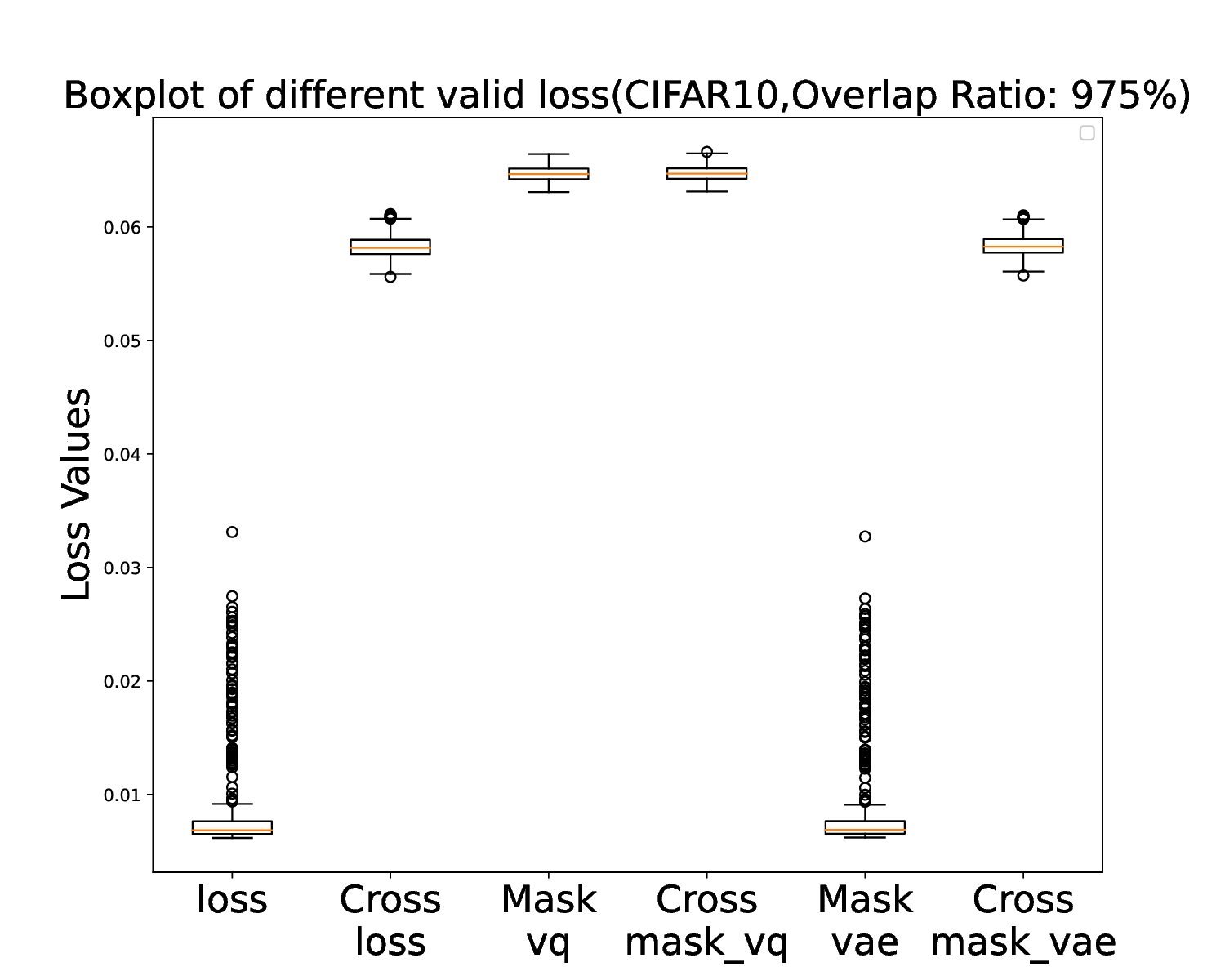}
    \setlength{\abovecaptionskip}{-0.05cm} %调整标题下方的距离
    \caption{The loss incurred by agents communicating in different ways. During the training process, a combination of continuous and discrete methods is used. During the validation phase, we employ three methods to obtain communication loss: the first way is to mask the content of the continuous information part and only use the discrete information part for communication; the second is to mask the content of the discrete information part and only use the continuous information part for communication; the third is to use both parts of information for integrated communication. When the familiarity between different pairs of agents exceeds 90\%, the effectiveness of communication using discrete semantics surpasses that of using continuous semantics. }
    % \vspace{-0.3cm}  %调整图片与上文的垂直距离
    \label{fig16}
\end{figure}
Figure \ref{fig16} represents some results when we trained using a combination of continuous and discrete methods, and used different validation methods during communication validation. The training method here involves splitting the encoder's output into two parts: one part goes through the latent variable layer of the VAE model, and the other part goes through the codebook layer of the VQ-VAE model, and then both parts are integrated into the decoder. During communication validation, we only use continuous, discrete, or a combination of continuous and discrete methods for cross-validation. The loss during the training process can be represented by Equation \ref{eq7}. 
\begin{equation}
\label{eq7}
\mathcal{L} = \mathcal{L} _{recon}+ \mathcal{L} _{quantization}  +  \mathcal{L}_{KL}         
\end{equation}
In this experiment, we found that when the overlap ratio is low, the effect of discrete communication is not as good as continuous communication. However, when the overlap ratio exceeds 90\%, agents learning through discrete communication overcome the problems in learning communication protocols, leading to a reduction in overall communication loss and outperforming continuous methods. However, the results of this experiment were obtained considering that both continuous and discrete information are present in the communication process of agents. When the agents learn and communicate entirely in a discrete or continuous manner, the advantage of the discrete method also disappears even with a 90\% overlap ratio.

Regarding the experiments on non-image datasets and the lack of demonstrated advantages over variational autoencoders, we will reserve them for continued research.

\bibliography{reference}

\begin{thebibliography}{10}

\bibitem{bib33}
David~H Ackley, Geoffrey~E Hinton, and Terrence~J Sejnowski.
\newblock A learning algorithm for boltzmann machines.
\newblock {\em Cognitive science}, 9(1):147--169, 1985.

\bibitem{bib22}
Anirudh~Goyal ALIAS PARTH~GOYAL, Aniket Didolkar, Nan~Rosemary Ke, Charles
  Blundell, Philippe Beaudoin, Nicolas Heess, Michael~C Mozer, and Yoshua
  Bengio.
\newblock Neural production systems.
\newblock {\em Advances in Neural Information Processing Systems},
  34:25673--25687, 2021.

\bibitem{bib15}
Diane Bouchacourt and Marco Baroni.
\newblock How agents see things: On visual representations in an emergent
  language game.
\newblock {\em arXiv preprint arXiv:1808.10696}, 2018.

\bibitem{bib5}
Rahma Chaabouni, Eugene Kharitonov, Emmanuel Dupoux, and Marco Baroni.
\newblock Anti-efficient encoding in emergent communication.
\newblock {\em Advances in Neural Information Processing Systems}, 32, 2019.

\bibitem{bib7}
Rahma Chaabouni, Eugene Kharitonov, Emmanuel Dupoux, and Marco Baroni.
\newblock Communicating artificial neural networks develop efficient
  color-naming systems.
\newblock {\em Proceedings of the National Academy of Sciences},
  118(12):e2016569118, 2021.

\bibitem{bib14}
Rahma Chaabouni, Florian Strub, Florent Altche, Eugene Tarassov, Corentin
  Tallec, Elnaz Davoodi, Kory~Wallace Mathewson, Olivier Tieleman, Angeliki
  Lazaridou, and Bilal Piot.
\newblock Emergent communication at scale.
\newblock In {\em International conference on learning representations}, 2021.

\bibitem{bib11}
Jakob Foerster, Ioannis~Alexandros Assael, Nando De~Freitas, and Shimon
  Whiteson.
\newblock Learning to communicate with deep multi-agent reinforcement learning.
\newblock {\em Advances in neural information processing systems}, 29, 2016.

\bibitem{bib31}
Washington Garcia, Hamilton Clouse, and Kevin Butler.
\newblock Disentangling categorization in multi-agent emergent communication.
\newblock In {\em Proceedings of the 2022 Conference of the North American
  Chapter of the Association for Computational Linguistics: Human Language
  Technologies}, pages 4523--4540, 2022.

\bibitem{bib16}
Allen Gersho and Robert~M. Gray.
\newblock Vector quantization and signal compression.
\newblock In {\em The Kluwer International Series in Engineering and Computer
  Science}, 1991.

\bibitem{bib23}
Anirudh Goyal, Aniket Didolkar, Alex Lamb, Kartikeya Badola, Nan~Rosemary Ke,
  Nasim Rahaman, Jonathan Binas, Charles Blundell, Michael Mozer, and Yoshua
  Bengio.
\newblock Coordination among neural modules through a shared global workspace.
\newblock {\em arXiv preprint arXiv:2103.01197}, 2021.

\bibitem{bib21}
Anirudh Goyal, Alex Lamb, Jordan Hoffmann, Shagun Sodhani, Sergey Levine,
  Yoshua Bengio, and Bernhard Scholkopf.
\newblock Recurrent independent mechanisms.
\newblock {\em arXiv preprint arXiv:1909.10893}, 2019.

\bibitem{bib27}
Shangmin Guo, Yi~Ren, Serhii Havrylov, Stella Frank, Ivan Titov, and Kenny
  Smith.
\newblock The emergence of compositional languages for numeric concepts through
  iterated learning in neural agents.
\newblock {\em arXiv preprint arXiv:1910.05291}, 2019.

\bibitem{bib32}
Serhii Havrylov and Ivan Titov.
\newblock Emergence of language with multi-agent games: Learning to communicate
  with sequences of symbols.
\newblock {\em Advances in neural information processing systems}, 30, 2017.

\bibitem{bib28}
Sepp Hochreiter and Jurgen Schmidhuber.
\newblock Long short-term memory.
\newblock {\em Neural computation}, 9(8):1735--1780, 1997.

\bibitem{bib29}
Yair Lakretz, German Kruszewski, Theo Desbordes, Dieuwke Hupkes, Stanislas
  Dehaene, and Marco Baroni.
\newblock The emergence of number and syntax units in lstm language models.
\newblock {\em arXiv preprint arXiv:1903.07435}, 2019.

\bibitem{bib25}
Alex Lamb, Di~He, Anirudh Goyal, Guolin Ke, Chien-Feng Liao, Mirco Ravanelli,
  and Yoshua Bengio.
\newblock Transformers with competitive ensembles of independent mechanisms.
\newblock {\em arXiv preprint arXiv:2103.00336}, 2021.

\bibitem{bib8}
Angeliki Lazaridou, Alexander Peysakhovich, and Marco Baroni.
\newblock Multi-agent cooperation and the emergence of (natural) language.
\newblock {\em arXiv preprint arXiv:1612.07182}, 2016.

\bibitem{bib1}
David Lewis.
\newblock Convention harvard university press.
\newblock {\em Cambridge, MA}, 1969.

\bibitem{bib34}
Sheng Li, Yutai Zhou, Ross Allen, and Mykel~J Kochenderfer.
\newblock Learning emergent discrete message communication for cooperative
  reinforcement learning.
\newblock In {\em 2022 International Conference on Robotics and Automation
  (ICRA)}, pages 5511--5517. IEEE, 2022.

\bibitem{bib20}
Dianbo Liu, Alex~M Lamb, Kenji Kawaguchi, Anirudh~Goyal ALIAS PARTH~GOYAL, Chen
  Sun, Michael~C Mozer, and Yoshua Bengio.
\newblock Discrete-valued neural communication.
\newblock {\em Advances in Neural Information Processing Systems},
  34:2109--2121, 2021.

\bibitem{bib9}
Ryan Lowe, Abhinav Gupta, Jakob Foerster, Douwe Kiela, and Joelle Pineau.
\newblock On the interaction between supervision and self-play in emergent
  communication.
\newblock {\em arXiv preprint arXiv:2002.01093}, 2020.

\bibitem{bib18}
Chris~J Maddison, Andriy Mnih, and Yee~Whye Teh.
\newblock The concrete distribution: A continuous relaxation of discrete random
  variables.
\newblock {\em arXiv preprint arXiv:1611.00712}, 2016.

\bibitem{bib30}
Hua Miao and Nanxiang Yu.
\newblock Targeted multi-agent communication with deep metric learning.
\newblock {\em Engineering Letters}, 31(2), 2023.

\bibitem{bib12}
Yaru Niu, Rohan~R Paleja, and Matthew~C Gombolay.
\newblock Multi-agent graph-attention communication and teaming.
\newblock In {\em AAMAS}, pages 964--973, 2021.

\bibitem{bib26}
Emanuele Pesce and G.~Montana.
\newblock Improving coordination in small-scale multi-agent deep reinforcement
  learning through memory-driven communication.
\newblock {\em Machine Learning}, 109:1727 -- 1747, 2019.

\bibitem{bib13}
Cinjon Resnick, Abhinav Gupta, Jakob Foerster, Andrew~M Dai, and Kyunghyun Cho.
\newblock Capacity, bandwidth, and compositionality in emergent language
  learning.
\newblock {\em arXiv preprint arXiv:1910.11424}, 2019.

\bibitem{bib17}
Jason~Tyler Rolfe.
\newblock Discrete variational autoencoders.
\newblock {\em arXiv preprint arXiv:1609.02200}, 2016.

\bibitem{bib10}
David Simoes, Nuno Lau, and Reis.
\newblock Multi-agent actor centralized-critic with communication.
\newblock {\em Neurocomputing}, 390:40--56, 2020.

\bibitem{bib35}
Olivier Tieleman, Angeliki Lazaridou, Shibl Mourad, Charles Blundell, and Doina
  Precup.
\newblock Shaping representations through communication: community size effect
  in artificial learning systems.
\newblock {\em arXiv preprint arXiv:1912.06208}, 2019.

\bibitem{bib19}
Aaron Van Den~Oord, Oriol Vinyals, et~al.
\newblock Neural discrete representation learning.
\newblock {\em Advances in neural information processing systems}, 30, 2017.

\bibitem{bib24}
Ashish Vaswani, Noam Shazeer, Niki Parmar, Jakob Uszkoreit, Llion Jones,
  Aidan~N Gomez, Lukasz Kaiser, and Illia Polosukhin.
\newblock Attention is all you need.
\newblock {\em Advances in neural information processing systems}, 30, 2017.

\end{thebibliography}
\bibliographystyle{plain}

\end{document}